\documentclass{article}

\usepackage[final]{neurips_2021}

\usepackage[utf8]{inputenc} 
\usepackage[T1]{fontenc}    
\usepackage{url}            
\usepackage{amsfonts}       
\usepackage{nicefrac}       
\usepackage{microtype}      
\usepackage{xcolor}         

\usepackage{rotating,amsmath,amssymb,amsfonts,amsthm}
\usepackage{epsfig,mathrsfs,natbib,color,graphics}
\usepackage{graphicx,algorithm,algorithmic}
\usepackage{array,wrapfig}
\usepackage{hyperref}
\usepackage{booktabs}
\usepackage{subcaption,adjustbox}
\usepackage{bbm}

\newcommand{\bx}{{\boldsymbol x}}

\newcommand{\bb}{{\boldsymbol b}}
\newcommand{\bw}{{\boldsymbol w}}

\newcommand{\bV}{{\boldsymbol V}}

\newcommand{\bt}{{\boldsymbol t}}

\newcommand{\mG}{{\cal G}}

\newcommand{\bTheta}{{\boldsymbol \Theta}}

\newcommand{\bbeta}{{\boldsymbol \beta}}
\newcommand{\bgamma}{{\boldsymbol \gamma}}

\newtheorem{theorem}{Theorem}[section]
\newtheorem{lemma}{Lemma}[section]
\newtheorem{definition}{Definition}[section]

\newtheorem{assumption}{Assumption}[section]
\title{Sparse Deep Learning: A New Framework Immune to Local Traps and Miscalibration}

\author{%
  Yan Sun  \\
  Purdue University\\
  West Lafayette, IN 47906 \\
  \texttt{sun748@purdue.edu} \\
  \And
  Wenjun Xiong \\
  Guangxi Normal University \& Purdue University \\
  West Lafayette, IN 47906 \\
  \texttt{xiong90@purdue.edu} \\
  \AND
  Faming Liang\thanks{To whom correspondence should be addressed: Faming Liang} \\
  Purdue University \\
  West Lafayette, IN 47906 \\
  \texttt{fmliang@purdue.edu} \\
}

\begin{document}

\maketitle

\begin{abstract}
Deep learning has powered recent successes of artificial intelligence (AI). However, the deep neural network, as the basic model of deep learning,  has suffered from issues such as local traps and miscalibration. In this paper, we provide a new framework for sparse deep learning, which has the above issues addressed in a coherent way. In particular, we lay down a theoretical foundation for sparse deep learning and propose prior annealing algorithms for learning sparse neural networks. 
The former has successfully tamed the sparse deep neural network into the framework of statistical modeling, enabling prediction uncertainty correctly quantified.
 The latter can be asymptotically guaranteed to converge to the global optimum, enabling the validity of the down-stream statistical inference. 
Numerical result indicates the superiority of the proposed method compared to the existing ones. 

\end{abstract}

\section{Introduction}

During the past decade, deep neural networks (DNNs) have achieved the state-of-the-art performance in many machine learning tasks such as computer vision and natural language processing. However, the DNN suffers from a training-prediction dilemma from the perspective of statistical inference: {\it A small DNN model can be well calibrated, but tends to get trapped into a local optimum; on the other hand, an over-parameterized DNN model can be easily trained to a global optimum (with zero training loss), but tends to be miscalibrated \citep{CalibrationDNN2017}}. 
In consequence, it is often unclear whether a DNN is guaranteed to have a desired property after training instead of getting trapped into an arbitrarily poor local minimum, or whether its decision/prediction is reliable. This difficulty  makes the trustworthiness of AI highly questionable.

To resolve this difficulty, researchers have attempted from two sides of the training-prediction dilemma. 
Towards understanding the optimization process of the DNN training, a line of researches have been done.
For example, \cite{GoriTesi1992} and \cite{Nguyen2017TheLS} studied the training loss surface of over-parameterized DNNs. They showed that for a fully connected DNN, almost all local minima are globally optimal, if the width of one layer of the DNN is no smaller than the training sample size and the network structure from this layer on is pyramidal. Recently, \cite{AllenZhu2019ACT,Du2019GradientDF, Zou2020GradientDO} and \cite{ZouGu2019} explored the convergence theory of the gradient-based algorithms in training over-parameterized DNNs. They showed that the gradient-based algorithms with random initialization can converge to global minima provided that the width of the DNN is polynomial in training sample size. 

To improve calibration of the DNN, different  methods have been developed, see e.g., Monte Carlo dropout \citep{MCdropout2016} and deep ensemble \citep{Deepensemble2017}. However, these methods did not provide a rigorous study for the asymptotic distribution of the DNN prediction and thus could not correctly quantify its uncertainty.
Recently, researchers have attempted to address this issue with sparse deep learning. For example, for Bayesian sparse neural networks, \cite{liang2018bayesian},   \cite{polson2018posterior} and \cite{SunSLiang2021} established the posterior consistency, and \cite{Wang2020UncertaintyQF} further established the Bernstein-von Mises (BvM) theorem for linear and quadratic functionals. The latter guarantees in theory that the Bayesian credible region has a faithful frequentist coverage. However, 
since the theory by \cite{Wang2020UncertaintyQF} does not cover
the point evaluation functional, the uncertainty of 
the DNN prediction still cannot be correctly quantified. 
Moreover, their theory is developed with the spike-and-slab prior (i.e., each weight or bias of the DNN is subject to a spike-and-slab prior), whose discrete nature makes the resulting posterior distribution extremely hard to simulate.  
To facilitate computation, \cite{SunSLiang2021} employed a mixture Gaussian prior. However, due to nonconvexity of the loss function of the DNN, a direct MCMC simulation still cannot be guaranteed to converge to the right posterior distribution even with the mixture Gaussian prior.

In this paper, we provide a new framework for sparse deep learning, which successfully resolved the training-prediction dilemma. In particular, we propose two prior annealing algorithms, one from the frequentist perspective and one from the Bayesian perspective, for learning sparse neural networks. The algorithms start with an over-parameterized deep neural network and then have its structure gradually sparsified. We provide a theoretical guarantee that the training procedures are immune to local traps, the resulting sparse structures are consistent, and the predicted values are asymptotically normally distributed. The latter enables the prediction uncertainty correctly quantified.  Our contribution in this paper is two-fold:

\begin{itemize}
    \item We provide a new framework for sparse deep learning, which is immune to local traps and miscalibration. 
    \item We lay down the theoretical foundation for how to make statistical inference with sparse deep neural networks. 
\end{itemize}

The remaining part of the paper is organized as follows. Section 2 lays down the theoretical foundation for sparse deep learning. Section 3 describes the proposed prior annealing algorithms. 
Section 4 presents some numerical results. Section 5 concludes the paper. 

 \section{Theoretical Foundation for Sparse Deep Learning}
   
As mentioned previously, sparse deep learning has received much attention as a promising way for addressing the miscalibration issue of the DNN. Theoretically, the approximation power of the sparse DNN has been studied for various classes of functions \citep{Schmidt-Hieber2017Nonparametric,Bolcskei2019}. Under the Bayesian setting, posterior consistency has been established in \cite{liang2018bayesian,polson2018posterior,SunSLiang2021}. In particular, the work \cite{SunSLiang2021} has achieved important progress toward taming sparse DNNs into the framework of statistical modeling. They provide a neural network approximation theory fundamentally different from the existing ones. In the existing theory, no data is involved and a small network can potentially achieve an arbitrarily small approximation error by allowing  connection weights to take values in an unbounded space\cite{maiorov1999lower}. In contrast, the theory by \cite{SunSLiang2021} links the network approximation error, the network size, and the bound of connection weights to the training sample size. They prove that 
  for a given training sample size $n$, a sparse DNN of size $O(n/\log(n))$ has been large enough to approximate many types of functions, such as affine functions and piecewise smooth functions, arbitrarily well as 
   $n\to \infty$. Moreover, they prove that the sparse DNN possesses many theoretical guarantees. For example, its structure is more interpretable, from which the relevant variables can be consistently identified for high-dimensional nonlinear systems; and its generalization error bound is asymptotically optimal. 
   
 From the perspective of statistical inference, some gaps remain toward taming sparse DNNs into the framework of statistical modeling. This paper bridges the gap by establishing (i) asymptotic normality of the connection weights, and  (ii) asymptotic normality of the prediction.
 
\subsection{Posterior consistency and structure selection consistency} 
  
 This subsection provides a brief review of the sparse DNN theory developed in \cite{SunSLiang2021} and gives the conditions that we will use in the followed theoretical developments. 
 Without loss of generality, we let 
$D_{n}=(\boldsymbol{x}^{(i)},y^{(i)})_{i=1,...,n}$ denote a dataset of $n$
$i.i.d$ observations, where $\boldsymbol{x}^{(i)}\in R^{p_n}$ and 
$y^{(i)}\in R$. Consider a generalized linear model with the distribution of $y$ given by
\[
f(y|\mu^*(\bx))=\exp\{A(\mu^*(\bx))y+B(\mu^*(\bx))+C(y)\},
\]
where $\mu^*(\bx)$ is a nonlinear function of $\bx$ and $A(\cdot)$,
 $B(\cdot)$ and $C(\cdot)$ are appropriately defined 
functions. For example, for normal regression, we have $A(\mu^*)=\mu^*/\sigma^2$, $B(\mu^*)=-{\mu^*}^2/2\sigma^2$, $C(y)=-y^2/2\sigma^2-\log(2\pi \sigma^2)/2$, and $\sigma^2$ is a constant.
We approximate $\mu^*(\bx)$ using a fully connected DNN with $H_{n}-1$ hidden layers. Let $L_h$ denote the number of hidden units at layer $h$ with $L_{H_n}=1$ for the output layer and $L_{0}=p_{n}$ for the input layer.  
Let  $\boldsymbol{w}^{h}\in \mathbb{R}^{L_{h}\times L_{h-1}}$ and $\boldsymbol{b}^{h}\in \mathbb{R}^{L_{h}\times1}$,
$h\in\{1,2,...,H_n\}$ denote the weights and bias of  layer $h$, and let 
 $\psi^{h}: \mathbb{R}^{L_{h}\times1}\to \mathbb{R}^{L_{h}\times1}$ denote a coordinate-wise
and piecewise differentiable activation function of layer $h$.
The DNN forms a nonlinear mapping
\begin{equation} \label{appeq}
\mu(\bbeta,\boldsymbol{x})=\boldsymbol{w}^{H_{n}}\psi^{H_{n}-1}\left[\cdots\psi^{1}\left[\boldsymbol{w}^{1}\boldsymbol{x}+\boldsymbol{b}^{1}\right]\cdots\right]+\boldsymbol{b}^{H_{n}},
\end{equation}
where $\bbeta=(\bw,\bb)=\big\{ {\bw}_{ij}^{h},{\bb}_{k}^{h} : h\in\{1,\ldots,H_{n}\}, i,k\in\{ 1,\ldots,L_{h}\}, j\in\{ 1,\ldots,L_{h-1}\} \big\}$
denotes the collection of all weights and biases,  consisting of $K_{n}=\sum_{h=1}^{H_{n}}\left(L_{h-1}\times L_{h}+L_{h}\right)$ elements. 
For convenience, we treat bias as a special connection and call each element in $\bbeta$ a connection weight. In order to represent the structure for a sparse DNN, we introduce an indicator variable for each connection weight. Let $\bgamma^{\boldsymbol{w}^{h}}$ and $\bgamma^{\boldsymbol{b}^{h}}$ 
denote the indicator variables associated with $\boldsymbol{w}^{h}$ and  $\boldsymbol{b}^{h}$, respectively. 
Let $\boldsymbol{\gamma}=\{ \bgamma_{ij}^{\boldsymbol{w}^{h}},\bgamma_{k}^{\boldsymbol{b}^{h}}:
 h\in\{1,\ldots,H_{n}\}$, $i,k\in\left\{ 1,\ldots,L_{h}\right\} ,j\in\left\{ 1,\ldots,L_{h-1}\right\}\} $, which specifies the structure of the sparse DNN. With slight abuse of notation, we will also write $\mu(\bbeta,\bx)$ as $\mu(\bbeta, \bx, \bgamma)$ to include the information of the network structure. We assume $\mu^*(\bx)$ can be well approximated 
by a {\it parsimonious neural network} with relevant variables, and call this parsimonious network 
as the {\it true DNN model}. More precisely,   
we define the {\it true DNN model} as 
\begin{equation} \label{trueDNNeq}
(\bbeta^*,\bgamma^*)=\underset{(\bbeta,\bgamma)\in \mG_n,\, \|\mu(\bbeta,\bgamma, \bx)-\mu^*(\bx)\|_{L^2(\Omega)} \leq \varpi_n}{\operatorname{arg\,min}}|\bgamma|, 
\end{equation}
where $\mG_n:=\mG(C_0,C_1,\varepsilon,p_n,H_n,L_1,L_2,\ldots,L_{H_n})$ denotes the space of valid sparse networks satisfying condition A.2 (given below) for the given values of $H_n$, $p_n$, and $L_h$'s,  and $\varpi_n$ is some sequence converging to 0 as $n \to \infty$. 
For any given DNN $(\bbeta,\bgamma)$, 
the error $\mu(\bbeta,\bgamma,\bx)-\mu^*(\bx)$ can be generally decomposed as the network approximation error $\mu(\bbeta^*,\bgamma^*,\bx)-\mu^*(\bx)$ and the
network estimation error $\mu(\bbeta,\bgamma,\bx)-\mu(\bbeta^*,\bgamma^*,\bx)$. The $L_2$ norm of the former is bounded by $\varpi_n$, and the order of the latter will be given in Lemma \ref{2normal}. For the sparse DNN, we make the following assumptions:

\begin{enumerate}
 \item[A.1] The input $\bx$ is bounded by 1 entry-wisely, i.e. $\bx\in \Omega=[-1,1]^{p_n}$, and the density of $\bx$ is bounded in its support $\Omega$ uniformly with respect to $n$.
 \item[A.2] The true sparse DNN model satisfies the following conditions:
 \begin{itemize}
 \item[A.2.1] The network structure satisfies:  
 $r_nH_n\log n$ $+r_n\log\overline L+s_n\log p_n\leq C_0n^{1-\varepsilon}$, where $0<\varepsilon<1$ is a small constant,
 $r_n=|\bgamma^*|$ denotes the connectivity of $\bgamma^*$, $\overline{L}=\max_{1\leq j\leq H_n-1}L_j$ denotes the maximum hidden layer width, 
 $s_n$ denotes the input dimension of $\bgamma^*$.
 
 \item[A.2.2] The network weights are polynomially bounded:  $\|\bbeta^*\|_\infty\leq E_n$, where 
  $E_n=n^{C_1}$ for some constant $C_1>0$.
 \end{itemize}
 \item[A.3] The activation function $\psi$ is Lipschitz continuous with a Lipschitz constant of 1.
 \end{enumerate}
 
 Refer to \cite{SunSLiang2021} for explanations and discussions on these assumptions.
 We let each connection weight and bias  be subject to a mixture Gaussian prior, i.e.,
\begin{equation} \label{marprior}
\begin{split}
& {w}_{ij}^{h}  \sim\lambda_{n}N(0,\sigma_{1,n}^{2})+(1-\lambda_{n})N(0,\sigma_{0,n}^{2}),  \quad  
 {b}_{k}^{h}  \sim\lambda_{n}N(0,\sigma_{1,n}^{2})+(1-\lambda_{n})N(0,\sigma_{0,n}^{2}),
\end{split}
\end{equation}
where $\lambda_n\in (0,1)$ is the mixture proportion, $\sigma_{0,n}^{2}$ is typically set to a very small number, while $\sigma_{1,n}^{2}$ is relatively large.

\paragraph{Posterior Consistency}
Let $P^{*}$ and
$E^{*}$ denote the respective probability measure and expectation with respect to data $D_{n}$.  
 Let $d(p_1,p_2)$ denote the Hellinger distance between two densities $p_{1}(\boldsymbol{x},y)$ and $p_{2}(\boldsymbol{x},y)$.
Let $\pi(A\mid D_{n})$ be the posterior probability of an event $A$. 
 
\begin{lemma}\label{2normal} (Theorem 2.1 of \cite{SunSLiang2021})
Suppose Assumptions A.1-A.3 hold. If the mixture Gaussian prior (\ref{marprior}) satisfies the conditions:
 $\lambda_n = O( 1/\{K_n[n^{H_n}(\overline Lp_n)]^{\tau}\})$ for some constant $\tau>0$,
  $E_n/\{H_n\log n+\log \overline L\}^{1/2}  \lesssim \sigma_{1,n} \lesssim n^{\alpha}$ for some 
  constant $\alpha>0$, and 
$\sigma_{0,n}  \lesssim \min\big\{ 1/\{\sqrt{n} K_n (n^{3/2} \sigma_{1,0} /H_n)^{H_n}\}$,  
 $1/\{\sqrt{n} K_n (n E_n/H_n)^{H_n}\} \big\}$, 
 then there exists an error sequence 
 $\epsilon_n^2 =O(\varpi_n^2)+O(\zeta_n^2)$ 
 such that $\lim_{n\to \infty} \epsilon_n= 0$ and  
 $\lim_{n\to \infty} n\epsilon_n^2= \infty$, and 
 the posterior distribution satisfies
 \begin{equation}\label{postcon}
\begin{split}
 & P^*\left\{ \pi[d(p_{\bbeta},p_{\mu^*}) > 4 \epsilon_n |D_n] \geq 2 e^{-cn \epsilon_n^2} \right\} 
  \leq 2 e^{-cn \epsilon_n^2},\\
 &  E_{D_n}^* \pi[d(p_{\bbeta},p_{\mu^*}) > 4 \epsilon_n | D_n] \leq 4 e^{-2cn \epsilon_n^2},
 \end{split}
\end{equation}
 for sufficiently large $n$, where $c$ denotes a constant,  $\zeta_n^2=[r_nH_n\log n+r_n\log \overline L+s_n\log p_n]/n$, $p_{\mu^*}$ denotes the underlying true data distribution, and $p_\bbeta$ denotes the data distribution reconstructed by the Bayesian DNN based on its posterior samples. 
\end{lemma}

\paragraph{Structure Selection Consistency}  The DNN is generally nonidentifiable due to the symmetry of network structure. 
For example, $\mu(\bbeta,\bgamma, \bx)$ can be invariant if one permutes certain hidden nodes or simultaneously changes the signs or scales of certain weights.
As in \cite{SunSLiang2021}, we define a set of DNNs by $\Theta$ such that any possible DNN can be represented by one and only one DNN in $\Theta$ via nodes permutation, sign changes, weight rescaling, etc.
Let $\nu(\bgamma,\bbeta) \in \Theta$ be an operator that maps any DNN to $\Theta$ via appropriate weight transformations. To serve the purpose of structure selection in $\Theta$, we consider the marginal inclusion posterior probability (MIPP) approach proposed in \cite{LiangSY2013}.
 For each connection, we define its MIPP by 
  $q_i=\int \sum_{\bgamma} e_{i|\nu(\bgamma,\bbeta)} \pi(\bgamma|\bbeta,D_n) 
   \pi(\bbeta|D_n) d\bbeta$ for $i=1,2,\ldots, K_n$,
 where $e_{i|\nu(\bgamma,\bbeta)}$ is the indicator 
 of connection $i$. 
 The MIPP approach is to choose the connections whose MIPPs are greater than a threshold $\hat{q}$, i.e., setting
 $\hat{\bgamma}_{\hat{q}}=\{i: q_i > \hat{q}, i=1,2,\ldots, K_n\}$ as
 an estimator of  
 $\bgamma^*\in \Theta$.
 Let $A(\epsilon_n)=\{\bbeta: d(p_\bbeta,p_{\mu^*})\geq \epsilon_n\}$ and
define
$
\rho(\epsilon_n) = \max_{1\leq i\leq K_n} \int_{A(\epsilon_n)^c}   \sum_{\bgamma} |e_{i|\nu(\bgamma,\bbeta)}-e_{i|\nu(\bgamma^*,\bbeta^*)}|\pi(\bgamma|\bbeta, D_n) 
 \pi(\bbeta|D_n)d\bbeta, 
$
 which measures the structure difference between the true and sampled models 
 on $A({\epsilon_n})^c$. Then we have:
 \begin{lemma} \label{Selectlem} (Theorem 2.2 of \cite{SunSLiang2021})  If the conditions of Lemma  \ref{2normal} hold and 
  $\rho(\epsilon_n) \to 0$ as $n\to \infty$ and $\epsilon_n\to 0$, then 
 (i)  $\max_{1\leq i\leq K_n}\{|q_i-e_{i|\nu(\bgamma^*,\bbeta^*)}|\}\stackrel{p}{\to} 0$; (ii) (sure screening)  
   $P(\bgamma_* \subset \hat{\bgamma}_{\hat{q}}) \stackrel{p}{\to} 1$ for any 
    pre-specified $\hat{q} \in (0,1)$;  
(iii) (consistency) $P(\bgamma_* =\hat{\bgamma}_{0.5}) \stackrel{p}{\to} 1$.
 \end{lemma}
 
 Lemma \ref{Selectlem} implies consistency of variable selection for the true DNN model as defined in (\ref{trueDNNeq}).

 \subsection{Asymptotic Normality of Connection Weights} 
In this section, we establish the asymptotic normality of the network parameters and predictions. 
 Let $n l_n(\bbeta) = \sum_{i=1}^n\log(p_{\bbeta}(\bx_i, y_i))$ denote the log-likelihood function, and let $\pi(\bbeta)$ denote the density of the mixture Gaussian prior (\ref{marprior}). Let $h_{i_{1},i_{2},\dots,i_{d}}(\bbeta)$
 denote the $d$-th order partial derivatives $\frac{\partial^{d} l_n(\bbeta)}{\partial\bbeta_{i_{1}}\partial\bbeta_{i_{2}}\cdots\partial\bbeta_{i_{d}}}$. Let $H_{n}(\bbeta)$ denote the Hessian matrix of $l_{n}(\bbeta)$. Let
 $h_{ij}(\bbeta)$ and $h^{ij}(\bbeta)$ denote the $(i,j)$-th component of $H_n(\bbeta)$ and $H_n^{-1}(\bbeta)$, respectively. 
 Let $\bar{\lambda}_n(\bbeta)$ and 
  $\underline{\lambda}_n(\bbeta)$ denotes the maximum and minimum eigenvalue of the Hessian matrix $H_n(\bbeta)$, respectively. Let $B_{\lambda,n}= {\bar{\lambda}_n^{1/2}(\bbeta^*)}/{ 
 \underline{\lambda}_n(\bbeta^*)}$ and 
  $b_{\lambda,n}=\sqrt{r_n/n} B_{\lambda,n}$, where $r_n$ is the connectivity of $\bgamma^*$.
 For a DNN parameterized by $\bbeta$, we define the weight truncation at the true model $\bgamma^*$: $(\bbeta_{\bgamma^*})_i = \bbeta_i$ for $i\in \bgamma^*$ and $(\bbeta_{\bgamma^*})_i = 0$ otherwise. For the mixture Gaussian prior (\ref{marprior}), let $B_{\delta_n}(\bbeta^*) = \{\bbeta: |\bbeta_i-\bbeta^*_i| < \delta_n, \forall i\in \bgamma^*, |\bbeta_i-\bbeta^*_i| < 2\sigma_{0,n}\log(\frac{\sigma_{1,n}}{\lambda_n\sigma_{0,n}}),\forall i \notin \bgamma^* \}$. 
We follow the definition of asymptotic normality in \cite{castillo2015bernstein} and \cite{Wang2020UncertaintyQF}:

\begin{definition}
Denote by  $d_{\bbeta}$ the bounded Lipschitz metric for weak convergence and  by $\phi_n$ the mapping $\phi_n: \bbeta \to \sqrt n (g(\bbeta)-g_*)$.
We say that the posterior distribution of the functional $g(\bbeta)$ is asymptotically normal with the center $g_*$ and variance $G$ if 
$d_{\bbeta}(\pi[\cdot\mid D_n]\circ \phi_n^{-1}, N(0, G))\to 0$
in ${P^*}$-probability  as $n\to \infty$. We  will write  this more compactly as $\pi[\cdot\mid D_n]\circ \phi_n^{-1} \rightsquigarrow N(0, G)$. 
\end{definition}
 
  Theorem \ref{bvm} establishes the asymptotic normality of $\tilde{\nu}(\bbeta)$, where $\tilde{\nu}(\bbeta)$ denotes a transformation of $\bbeta$ which is invariant with respect to   $\mu(\bbeta,\bgamma,\bx)$ while minimizing $\|\tilde{\nu}(\bbeta)-\bbeta^*\|_{\infty}$. 
 
 \begin{theorem}\label{bvm}
 Assume the conditions of Lemma \ref{Selectlem} hold with $\rho(\epsilon_n) = o(\frac{1}{K_n})$ and $C_1>\frac{2}{3}$ in Condition A.2.2.
For some $\delta_n$ s.t. $\frac{r_n}{\sqrt{n}} \lesssim \delta_n \lesssim \frac{1}{\sqrt[3]{n}r_n}$, let $A(\epsilon_n,\delta_n) = \{\bbeta:\max_{i\in \bgamma^{*}}|\bbeta_{i} - \bbeta_i^*| > \delta_n,  d(p_\bbeta,p_{\mu^*}) \leq \epsilon_n\}$, where $\epsilon_n$ is the posterior contraction rate 
as defined in Lemma \ref{2normal}.
Assume there exists some constants $C>2$ and $M>0$ 
such that 
\begin{enumerate}
\item[C.1] $\bbeta^*=(\bbeta_1^*,\bbeta_2^*,\ldots,\bbeta_{K_n}^*)$ is generic \cite{feng2017sparse, fefferman1994reconstructing}, $\min_{i\in \bgamma^*}|\bbeta^*_i| > C\delta_n$ and $\pi(A(\epsilon_n,\delta_n) \mid D_n) \rightarrow 0$ as $n\rightarrow \infty$.
\item [C.2] $|h_i(\bbeta^*)|<M$, $|h_{j,k}(\bbeta^*)| < M$, $|h^{j,k}(\bbeta^*)| < M$, $|h_{i,j,k}(\bbeta)| < M$, $|h_l(\bbeta)|<M$ hold for any $i,j,k \in \bgamma^*$, $l \notin \bgamma^*$ and $\bbeta\in B_{2\delta_n}(\bbeta^*)$.
 \item [C.3] 
 $\sup\left\{ |E_{\bbeta}(a^T U)^3|: \|\bbeta_{\gamma^*}-\bbeta^*\| \leq 1.2 b_{\lambda,n}, \|a\|=1 \right\} \leq 0.1 \sqrt{n/r_n} \underline{\lambda}_n^2(\bbeta^*)/\bar{\lambda}_n^{1/2}(\bbeta^*)$ and $B_{\lambda,n}=O(1)$, where $U=Z-E_{\bbeta_{\gamma^*}} (Z)$,
 $Z$ denotes a random variable drawn from a neural network model parameterized by $\bbeta_{\gamma^*}$, and $E_{\bbeta_{\gamma^*}}(Z)$ denotes the mean of $Z$. 
\end{enumerate}
Then $\pi[\sqrt n (\tilde{\nu}(\bbeta)-\bbeta^*) \mid D_n] \rightsquigarrow N(0, \bV)$ 
in ${P^*}$-probability  as $n\to \infty$, where $\bV=(v_{ij})$, and  $v_{i,j} = E(h^{i,j}(\bbeta^*))$ if $i,j \in \bgamma^*$ and $0$ otherwise.
\end{theorem}

Condition C.1 is essentially an identifiability condition, i.e., when $n$ is sufficiently large, 
the DNN weights cannot be too far away from the true weights if the DNN produces approximately the same  distribution as the true data.  Condition C.2 gives typical conditions on derivatives of the DNN. Condition C.3 ensures consistency of the MLE  of $\bbeta^*$
for the given structure $\bgamma^*$ \cite{Portnoy1988}.  

\subsubsection{Asymptotic Normality of Prediction}

Theorem \ref{predictionbvm} establishes asymptotic normality of the prediction $\mu(\bbeta, \bx_0)$ for a test data point $\bx_0$, which implies that a faithful prediction interval can be constructed for the learnt sparse neural network. Refer to Appendix A.4 for how to construct the prediction interval based on the theorem.
Let $\mu_{i_{1},i_{2},\dots,i_{d}}(\bbeta,\bx_0)$ denote the $d$-th order partial derivative   $\frac{\partial^{d}\mu(\bbeta,\bx_0)}{\partial\bbeta_{i_{1}}\partial\bbeta_{i_{2}}\cdots\partial\bbeta_{i_{d}}}$.

\begin{theorem}\label{predictionbvm} Assume the conditions of Theorem \ref{bvm} and the following condition hold:  $|\mu_i(\bbeta^*,\bx_0)| < M$, $|\mu_{i,j}(\bbeta,\bx_0)| < M$, $|\mu_k(\bbeta,\bx_0)| < M$ hold for any $i,j \in \bgamma^*, k \notin \bgamma^*$ and $\bbeta\in B_{2\delta_n}(\bbeta^*)$, 
where $M$ is as defined in Theorem \ref{bvm}. 
Then   $\pi[\sqrt n (\mu(\bbeta, \bx_0)-\mu(\bbeta^*,\bx_0)) \mid D_n] \rightsquigarrow N(0, \Sigma)$,
where $\Sigma = \nabla_{\bgamma^*}\mu(\bbeta^*, \bx_0)^{T}H^{-1}\nabla_{\bgamma^*}\mu(\bbeta^*, \bx_0)$
 and  $H = E(-\nabla^2_{\bgamma^*} l_n(\bbeta^*))$ is the Fisher information matrix.
\end{theorem}

  The asymptotic normality for general smooth functional has been established in \cite{castillo2015bernstein}. 
  For linear and quadratic functional of deep ReLU network with a spike-and-slab prior, the 
  asymptotic normality has been established in \cite{Wang2020UncertaintyQF}. The DNN prediction $\mu(\bbeta, \bx_0)$ can be viewed as a point evaluation functional over the neural network function space. However, in general, this functional is not smooth with respect to the locally asymptotic normal (LAN) norm. The results of \cite{castillo2015bernstein} and \cite{Wang2020UncertaintyQF} are not directly applicable for the asymptotic normality of $\mu(\bbeta, \bx_0)$. 

\section{Prior Annealing Algorithms for Sparse DNN Computation}

As implied by Theorems \ref{bvm} and  \ref{predictionbvm}, a consistent estimator of $(\bgamma^*,\bbeta^*)$ is essential for statistical inference of the sparse DNN. 
Toward this goal, \cite{SunSLiang2021} proved that the marginal inclusion probabilities $q_i$'s can be estimated using Laplace approximation at the mode of the log-posterior.  Based on this result, they proposed a multiple-run procedure. In each run, they first maximize the log-posterior by an optimization algorithm, such as SGD or Adam; then sparsify the DNN structure by truncating the weights less than a threshold to zero, where the threshold is calculated from the prior (\ref{marprior}) based on the Laplace approximation theory; and then refine the weights of the sparsified DNN by running an optimization algorithm for a few iterations. Finally, they select a sparse DNN model from those obtained in the multiple runs according to their Bayesian evidence or BIC values. The BIC is suggested when the size of the sparse DNN is large. 

Although the multiple-run procedure works well for many problems, it is hard to justify that it will lead to a consistent estimator of the true model $(\bgamma^*,\bbeta^*)$. To tackle this issue, 
 we propose two prior annealing algorithms, one from the frequentist perspective and one from the Bayesian perspective.  

\subsection{Prior Annealing: Frequentist Computation} 

It has been shown in \cite{Nguyen2017TheLS, GoriTesi1992} that the loss of an over-parameterized DNN exhibits good properties:

\begin{itemize}
\item[($S^*$)]  For a fully connected DNN with an analytic activation function and a convex loss function at the output layer, if the number of hidden units of one layer is larger than the number of training points and the network structure from this layer on is pyramidal, then almost all local minima are globally optimal.
\end{itemize}
  
 Motivated by this result, we propose a prior annealing algorithm, which is immune to local traps and aims to find a consistent estimate of 
 $(\bbeta^*,\bgamma^*)$ as defined in (\ref{trueDNNeq}). 
 The detailed procedure of the algorithm is given in Algorithm \ref{priorannealing}. 
 
\begin{algorithm}  
\caption{Prior annealing: Frequentist} \label{priorannealing}
\begin{itemize}
\item[(i)]({\it Initial training}) Train a DNN 
 satisfying condition (S*) such that a global optimal solution $\bbeta_0=
\arg\max_{\bbeta} l_n(\bbeta)$ is reached, 
which can be accomplished using SGD or Adam \cite{adam2015}. 

\item[(ii)] ({\it Prior annealing}) Initialize $\bbeta$ at $\bbeta_0$ and simulate from 
a sequence of distributions  
 $\pi(\bbeta|D_n,\tau,\eta^{(k)}, \sigma_{0,n}^{(k)})$ $\propto e^{n l_n(\bbeta)/\tau} \pi_{k}^{\eta^{(k)}/\tau}(\bbeta)$ for $k=1,2,\ldots,m$, where $0<\eta^{(1)} \leq \eta^{(2)} \leq \cdots  \leq \eta^{(m)}=1$, 
 $\pi_k = \lambda_n N(0, \sigma_{1,n}^2) + (1-\lambda_n) N(0, (\sigma_{0,n}^{(k)})^{2})$, and $\sigma_{0,n}^{init} = \sigma_{0,n}^{(1)} \geq \sigma_{0,n}^{(2)} \geq \cdots \geq \sigma_{0,n}^{(m)} = \sigma_{0,n}^{end}$. The simulation can be done in an annealing manner using 
 a stochastic gradient MCMC algorithm  \citep{welling2011bayesian,SGHMC2014,ma2015complete,nemeth2019stochastic}. 
 After the stage $m$ has been reached, continue to run the simulated annealing algorithm by gradually decreasing the temperature $\tau$ to a very small value. Denote the resulting DNN by $\hat{\bbeta}=(\hat{\bbeta}_1,\hat{\bbeta}_2,\ldots,
  \hat{\bbeta}_{K_n})$.
 
\item[(iii)]  ({\it Structure sparsification}) For each connection $i \in \{1,2,\ldots,K_n\}$, 
 set $\tilde{\bgamma}_{i}=1$ if $|\hat{\bbeta}_i|>\frac{\sqrt{2} \sigma_{0,n}\sigma_{1,n}}{\sqrt{\sigma_{1,n}^2-\sigma_{0,n}^2}}$ $\sqrt{\log\left( \frac{1-\lambda_n}{\lambda_n}  \frac{\sigma_{1,n}}{\sigma_{0,n}} \right)}$ and 0 otherwise, 
 where the threshold value of $|\hat{\bbeta}_i|$ is obtained by solving 
 $\pi(\bgamma_i = 1 | \bbeta_i) > 0.5$ based on the mixture Gaussian prior as in \cite{SunSLiang2021}.
  Denote the yielded sparse DNN structure by $\tilde{\bgamma}$.  
  
 \item[(iv)] ({\it Nonzero-weights refining}) Refine the nonzero weights of the sparsified DNN by maximizing $l_n(\bbeta)$.
  Denote the resulting estimate by $\tilde{\bbeta}_{\tilde{\bgamma}}$, which represents 
   the MLE of $\bbeta^*$.
\end{itemize}
\end{algorithm}

For Algorithm \ref{priorannealing}, the consistency of $(\tilde{\bgamma}, \tilde{\bbeta}_{\tilde{\bgamma}})$ as an estimator of  $(\bgamma^*,\bbeta^*)$ can be proved based on Theorem 3.4 of \cite{Nguyen2017TheLS} for global convergence of $\bbeta_0$, the property of simulated annealing (by choosing an appropriate sequence of $\eta_k$ and a cooling schedule of $\tau$), Theorem \ref{Selectlem} for consistency of structure selection, Theorem 2.3 of \cite{SunSLiang2021} for consistency of structure sparsification, and 
Theorem 2.1 of \cite{Portnoy1988} for consistency of MLE under the scenario of dimension diverging.
Then we can construct the confidence intervals for neural network predictions using $(\tilde{\bgamma}, \tilde{\bbeta}_{\tilde{\bgamma}})$. The detailed procedure is given in supplementary material. 

Intuitively, the initial training phase can reach the global optimum of the likelihood function. In the prior annealing phase, as we slowly add the effect of the prior,  the landscape of the target distribution is gradually changed and the MCMC algorithm is likely to hit the region around the optimum of the target distribution. More explanations on the effect of the prior can be found in the supplementary material.
In practice, let $t$ denote the step index, a simple implementation of the initial training and prior annealing phases of Algorithm \ref{priorannealing} can be given as follows:
(i) for $0 < t < T_1$, run initial training; 
(ii) for $T_1 \leq t \leq T_2$, fix $\sigma_{0,n}^{(t)} = \sigma_{0,n}^{init}$ and linearly increase $\eta_t$ by setting $\eta^{(t)} = \frac{t - T_1}{T_2 - T_1}$; 
(iii) for $T_2 \leq t \leq T_3$, fix $\eta^{(t)} = 1$ and linearly decrease $\left(\sigma_{0,n}^{(t)}\right)^2$ by setting  $\left(\sigma_{0,n}^{(t)}\right)^2 = \frac{T_3 - t}{T_3 - T_2} \left(\sigma_{0,n}^{init}\right)^2 + \frac{t - T_2}{T_3 - T_2} \left(\sigma_{0,n}^{end}\right)^2$;
(iv) for $t > T_3$, fix $\eta^{(t)} = 1$ and $\sigma_{0,n}^{(t)} = \sigma_{0,n}^{end}$ and gradually decrease the temperature $\tau$, e.g., setting $\tau_t = \frac{c}{t-T_3}$ for some constant $c$.

\subsection{Prior Annealing: Bayesian Computation} 

For certain problems the size (or \#nonzero elements) of $\bgamma^*$ is large, calculation of the Fisher information matrix is difficult. In this case, the prediction uncertainty can be quantified via posterior simulations. The simulation can be started with a DNN  satisfying condition (S*) and performed 
using a SGMCMC algorithm \cite{ma2015complete, nemeth2019stochastic} with an annealed prior as defined in step (ii) of Algorithm \ref{priorannealing} (For Bayesian approach, we may fix the temperature $\tau=1$). 
The over-parameterized structure and 
annealed prior make the simulations immune to local traps.

To justify the Bayesian estimator for the prediction mean and variance, we study the deviation of the path averaging estimator $\frac{1}{T}\sum_{t=1}^{T}\phi(\bbeta^{(t)})$ and the posterior mean $\int \phi(\bbeta) \pi(\bbeta|D_n, \eta^*, \sigma_{0,n}^*) d\bbeta$ for some test function $\phi(\bbeta)$. For simplicity, we will focus on SGLD with prior annealing. Our analysis can be easily generalized to other SGMCMC algorithms \cite{chen2015convergence}.

For a test function $\phi(\cdot)$, the difference between $\phi(\bbeta)$ and $\int \phi(\bbeta) \pi(\bbeta|D_n, \eta^*, \sigma_{0,n}^*) d\bbeta$ can be characterized by the Poisson equation:
\[
\mathcal{L} \psi(\bbeta) = \phi(\bbeta) - \int \phi(\bbeta) \pi(\bbeta|D_n, \eta^*, \sigma_{0,n}^*) d\bbeta,
\]
where $\psi(\cdot)$ is the solution of the Poisson equation and $\mathcal{L}$ is the infinitesimal generator of the Langevin diffusion. i.e. for the following Langevin diffusion
$\text{d}\bbeta^{(t)} = \nabla\log(\pi(\bbeta|D_n, \eta^*, \sigma_{0,n}^*)) \text{d}t + \sqrt{2}I \text{d}W_t$,
where $I$ is identity matrix and $W_t$ is Brownian motion, we have 
\[
\mathcal{L}\psi(\bbeta) := \langle \nabla\psi(\bbeta), \nabla\log(\pi(\bbeta|D_n, \eta^*, \sigma_{0,n}^*)) + \text{tr}(\nabla^2 \psi(\bbeta)).
\]
Let $\mathcal{D}^k\psi$ denote the kth-order derivatives of $\psi$. To control the perturbation of $\phi(\bbeta)$, we need the following assumption about the function $\psi(\bbeta)$:

\begin{assumption}
\label{anneal_assump_1}
For $k\in \{0,1,2,3\}$, $\mathcal{D}^k\psi$ exists and there exists a function $\mathcal{V}$, s.t. $||\mathcal{D}^k \psi|| \lesssim \mathcal{V}^{p_k}$ for some constant $p_k > 0$. In addition, $\mathcal{V}$ is smooth and the expectation of $\mathcal{V}^p$ on $\bbeta^{(t)}$ is bounded for some $p\leq 2 \max_{k} \{p_k\}$, i.e. $\sup_{t} \mathbb{E} (\mathcal{V}^p(\bbeta^{(t)})) < \infty$, $\sum_{s\in(0,1)}\mathcal{V}^p(s\bbeta_1 + (1 - s) \bbeta_2) \lesssim \mathcal{V}^p(\bbeta_1) + \mathcal{V}^p(\bbeta_2)$. 
\end{assumption}

In step $t$ of the SGLD algorithm, the drift term is replaced by $\nabla_{\bbeta}\log\pi(\bbeta^{(t)}|D_{m,n}^{(t)}, \eta^{(t)}, \sigma_{0,n}^{(t)})$, where $D_{m,n}^{(t)}$ is used to represent the mini-batch data used in step t. Let $\mathcal{L}_t$ be the corresponding infinitesimal generator. Let $\delta_t = \mathcal{L}_t - \mathcal{L}$. To quantify the effect of $\delta_t$, we introduce the following assumption:

\begin{assumption}
\label{anneal_assump_2}
$\bbeta^{(t)}$ has bounded expectation and the expectation of log-prior is Lipschitz continuous with respect to $\sigma_{0,n}$, i.e. there exists some constant $M$ s.t. $\sup_{t}\mathbb{E}(|\bbeta^{(t)}|) \leq M < \infty$.   For all $t$, $|\mathbb{E} \log(\pi(\bbeta^{(t)}| \lambda_n, \sigma_{0,n}^{(t_1)}, \sigma_{1,n})) - \mathbb{E} \log(\pi(\bbeta^{(t)}| \lambda_n, \sigma_{0,n}^{(t_2)}, \sigma_{1,n}))| \leq M |\sigma_{0,n}^{(t_1)} - \sigma_{0,n}^{(t_2)}|$.
\end{assumption}
Then we have the following theorem:
\begin{theorem}
\label{anneal_thm}
Suppose the model satisfy assumption \ref{anneal_assump_2}, and a constant learning rate of $\epsilon$ is used. For a test function $\phi(\cdot)$, if the solution of the Poisson equation $\psi(\cdot)$ satisfy assumption \ref{anneal_assump_1}, then 
\begin{equation}
\small
\mathbb{E}\left(\frac{1}{T} \sum_{t = 1}^{T - 1} \phi(\bbeta^{(t)}) - \int \phi(\bbeta) \pi(\bbeta|D_n, \eta^*, \sigma_{0,n}^*) d\bbeta \right)
= O\left(\frac{1}{T\epsilon} + \frac{\sum_{t = 0}^{T-1} (|\eta^{(t)} - \eta^*| +|\sigma_{0,n}^{(t)} - \sigma_{0,n}^{*} |)}{T} + \epsilon\right),
\end{equation}
where $\sigma_{0,n}^*$ is treated as a fixed constant.
\end{theorem}
Theorem \ref{anneal_thm} shows that with prior annealing, the path averaging estimator can still be used for estimating the mean and variance of the  prediction and constructing the confidence interval. The detailed procedure is given in supplementary material. For the case that a decaying learning rate is used, a similar theorem can be developed as in \cite{chen2015convergence}.

\section{Numerical Experiments}

This section illustrates the performance of the proposed method on synthetic and real data examples.\footnote{The code for running these experiments can be found in \url{https://github.com/sylydya/Sparse-Deep-Learning-A-New-Framework-Immuneto-Local-Traps-and-Miscalibration}} 
For the synthetic example, the frequentist algorithm is employed to construct prediction intervals. The real data example involves a large network, so both the frequentist and Bayesian algorithms are employed along with comparisons with some existing network pruning methods.  

\subsection{Synthetic Example}

We consider a high-dimensional nonlinear regression problem, which shows that our method can identify the sparse network structure and relevant features as well as produce prediction intervals with correct coverage rates. The datasets were generated as in \cite{SunSLiang2021}, where the explanatory variables $x_{1},\dots,x_{p_n}$ were simulated  by independently generating $e,z_{1},\dots,z_{p_n}$ from $N(0,1)$ and setting $x_{i}=\frac{e+z_{i}}{\sqrt{2}}$. 
The response variable was generated from a nonlinear regression model:
\begin{equation}
\nonumber
\begin{split}
y&=\frac{5x_{2}}{1+x_{1}^{2}}+5\sin(x_{3}x_{4})+2x_{5} +0x_{6}+\cdots+0x_{2000}+\epsilon,
\end{split}
\end{equation}
where $\epsilon \sim N(0,1)$. 
Ten datasets were generated, each consisting of  10000 samples for training  and 1000 samples for testing.
 This example was taken from \cite{SunSLiang2021}, through which we show that the prior annealing method can achieve similar results with the multiple-run method proposed in \cite{SunSLiang2021}.

We modeled the data by a DNN of structure 2000-10000-100-10-1 with tanh activation function. Here we intentionally made the network very wide in one hidden layer to satisfy the condition (S*). 
Algorithm \ref{priorannealing} was employed to learn the model. The detailed setup for the experiments were given in the supplementary material.
The variable selection 
performance were measured using the false
selection rate 
$FSR=\frac{\sum_{i=1}^{10}|\hat{S}_{i}\backslash S|}{\sum_{i=1}^{10}|\hat{S}_{i}|}$ and 
negative selection rate 
$NSR=\frac{\sum_{i=1}^{10}|S\backslash\hat{S}_{i}|}{\sum_{i=1}^{10}|S|}$,
where $S$ is the set of true variables,
$\hat{S}_{i}$ is the set
of selected variables 
from dataset $i$ and $|\hat{S}_{i}|$ is the
size of $\hat{S}_{i}$. The predictive performance is measured by mean square prediction error (MSPE) and mean square fitting
error (MSFE). We compare our method with the 
multiple-run method (BNN\_evidence) \cite{SunSLiang2021}   as well as other existing variable selection methods including Sparse input neural network(Spinn) \cite{feng2017sparse},
Bayesian adaptive regression tree (BART) \cite{bleich2014variable},
linear model with lasso penalty (LASSO) \cite{tibshirani1996regression},
and sure independence screening with SCAD penalty (SIS)\cite{fan2008sure}. To demonstrate the importance of selecting correct variables, we also compare our method with two dense model with the same network structure: DNN trained with dropout(Dropout) and DNN trained with no regularization(DNN). Detailed setups for these methods were given in the supplementary material as well. 
The results were summarized in Table \ref{Simulation}. With a single run, our method BNN\_anneal achieves similar result with the multiple-run method. 
The latter trained the model for 10 times and selected the best one using Bayesian evidence.
While for Spinn (with LASSO penalty), even with over-parametrized structure, it performs worse than the sparse BNN model.

\begin{table*}[t]
\caption{Simulation Result: MSFE and MSPE were calculated by averaging over 10 datasets, and their standard deviations were given in the parentheses. }
\label{Simulation}
\begin{center}
\begin{tabular}{ccccccc}
 \toprule
Method & $|\hat S|$ & FSR & NSR & MSFE & MSPE \\
  \midrule
BNN\_anneal & 5(0) & 0 & 0 & 2.353(0.296) & 2.428(0.297) \\ 
BNN\_Evidence & 5(0) & 0 & 0 & 2.372(0.093) & 2.439(0.132) \\ 
Spinn & 10.7(3.874) & 0.462 & 0 & 4.157(0.219) & 4.488(0.350) \\
DNN & - & - & - & 1.1701e-5(1.1542e-6) & 16.9226(0.3230) \\
Dropout & - & - & - & 1.104(0.068) & 13.183(0.716) & \\
BART50 & 16.5(1.222) & 0.727 & 0.1 & 11.182(0.334) & 12.097(0.366) & \\
LASSO & 566.8(4.844) & 0.993 & 0.26 & 8.542(0.022) & 9.496(0.148) & \\
SIS & 467.2(11.776) & 0.991 & 0.2 & 7.083(0.023) & 10.114(0.161) & \\
 \bottomrule
\end{tabular}
\end{center}
\end{table*}

\begin{figure}
\centering
\includegraphics[height=2.5in,width=6.0in]{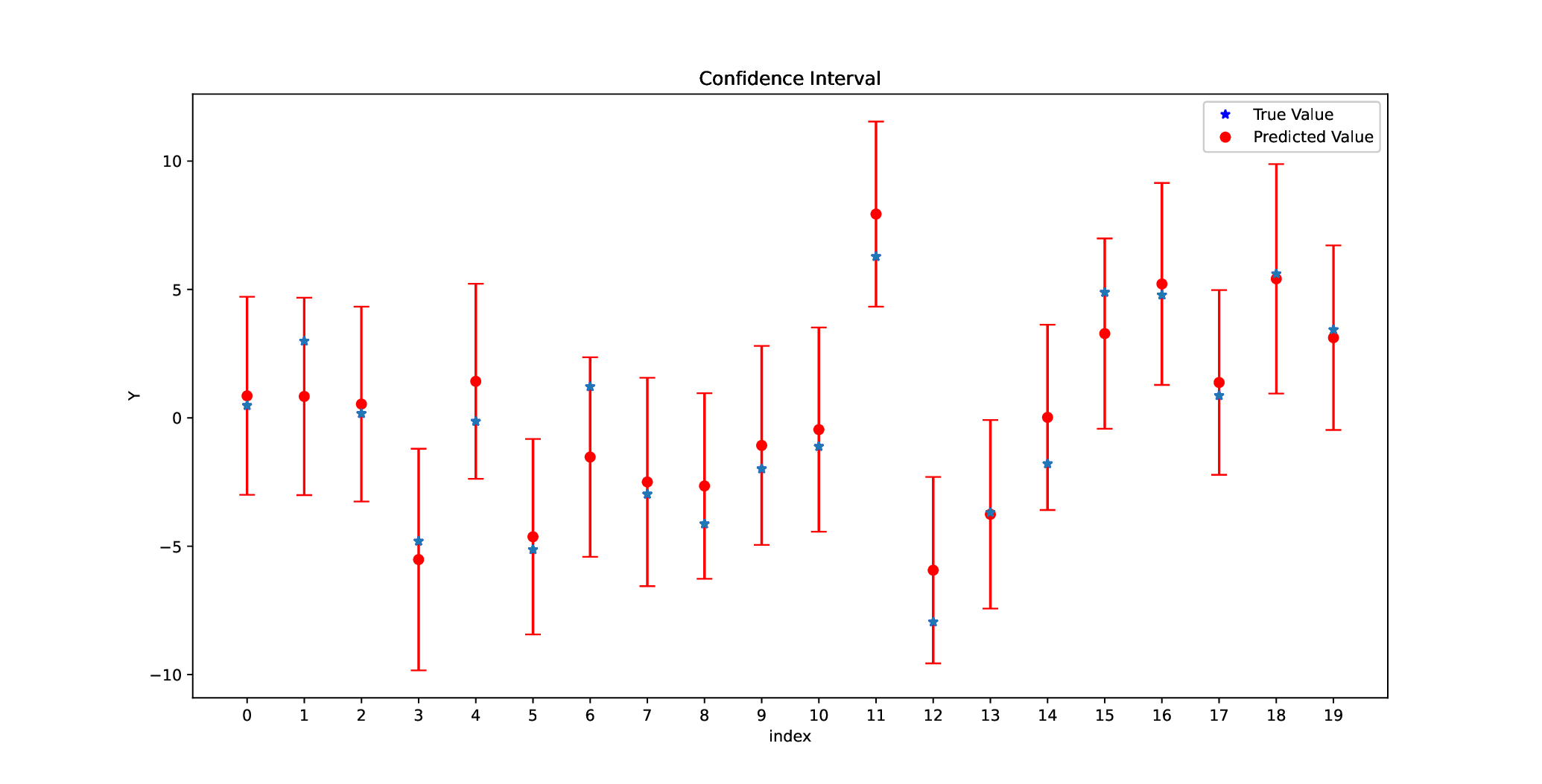}
\caption{Prediction intervals of 20 testing points, where the y-axis is the response value, the x-axis is the index, and the blue point represents the true observation.}
\label{CI}
\vspace{-0.1in}
\end{figure}

To quantify the uncertainty of the prediction, 
we conducted 100 experiments over different training sets as generated previously. We constructed 95\% prediction intervals over 1000 test points. Over the 1000 test points, the average coverage rate of the prediction intervals is $94.72\%(0.61\%)$, where $(0.61\%)$ denote the standard deviation. Figure \ref{CI} shows the prediction intervals constructed for 20 of the testing points. Refer to the supplementary material for the detail of the computation.  

\subsection{Real Data Example}

As a different type of applications of the proposed method, we conducted unstructured network pruning experiments on CIFAR10 dataset\cite{krizhevsky2009learning}.
Following the setup in \cite{lin2020dynamic}, we train the  residual network\cite{he2016deep} with different networks size and pruned the network to different sparsity levels.
The detailed experimental setup can be found in the  supplementary material. 

We compared the proposed methods, BNN\_anneal (Algorithm \ref{priorannealing}) and BNN\_average (averaged over last 75 networks simulated by the Bayesian version of the prior annealing algorithm), with several state-of-the-art unstructured pruning methods, including Consistent Sparse Deep Learning (BNN\_BIC) \cite{SunSLiang2021}, Dynamic pruning with feedback (DPF) \cite{lin2020dynamic}, Dynamic Sparse Reparameterization (DSR) \cite{mostafa2019parameter} and Sparse Momentum (SM) \cite{dettmers2019sparse}. The results of the baseline methods were taken from \cite{lin2020dynamic} and \cite{SunSLiang2021}. The results of prediction accuracy for different models and target sparsity levels were summarized in Table \ref{CIFAR}. Due to the threshold used in step (iii) of Algorithm \ref{priorannealing}, it is hard for our method to make the pruning ratio exactly the same as the targeted one. We intentionally make the pruning ratio smaller than the target ratio, while our method still achieve better test accuracy. Compared to BNN\_BIC, the test accuracy is very close, but the result of BNN\_BIC is obtained by running the experiment 10 times while our method only run once. To further demonstrate that the proposed method result in better model calibration, we followed the setup of \cite{maddox2019simple} and compared  
the proposed method with DPF on several metrics designed for  model calibration, including negtive log likelihood (NLL), symmetrized, discretized KL distance between in and out of sample entropy distributions (JS-Distance), and expected calibration error (ECE). For JS-Distance, we used the test data of SVHN data set\footnote{The Street View House Numbers (SVHN) Dataset: \url{http://ufldl.stanford.edu/housenumbers/}} as out-of-distribution samples. The results were summarized in Table \ref{CIFAR_calibration}. As discussed in \cite{maddox2019simple,CalibrationDNN2017}, a well calibrated model tends to have smaller NLL, larger JS-Distance and smaller ECE. The comparison shows that the proposed method outperforms DPF in most cases. In addition to the network pruning method, we also train a dense model with the standard training set up. Compared to the dense model, the sparse network has worse accuracy, but it tends to outperform the dense network in terms of ECE and JS-Distance, which indicates that sparsification is also a useful way for improving calibration of the DNN.

\begin{table}
\caption{ResNet network pruning results for CIFAR-10 data, which were calculated by averaging over 3 independent runs with the standard deviation reported in the parentheses. 
}
\label{CIFAR}
\begin{center}
\begin{tabular}{ccccccccc} \toprule
     & \multicolumn{2}{c}{ResNet-20} & & 
     \multicolumn{2}{c}{ResNet-32} \\ \cline{2-3}\cline{5-6}
Method &  Pruning Ratio & Test Accuracy & & Pruning Ratio &  Test Accuracy \\
  \midrule
DNN\_dense & 100\% & 92.93(0.04) & & 100\% & 93.76(0.02) \\
  \midrule
BNN\_average & 19.85\%(0.18\%) &  92.53(0.08) & & 9.99\%(0.08\%) & {93.12(0.09)} \\ 
BNN\_anneal  & 19.80\%(0.01\%) & {92.30(0.16)} & & 9.97\%(0.03\%) & {92.63(0.09)} \\ 
BNN\_BIC & 19.67\%(0.05\%) & {92.27(0.03)} & &  9.53\%(0.04\%) & {92.74(0.07)}  \\ 
SM  & 20\% & 91.54(0.16) & &  10\% & 91.54(0.18)   \\
DSR & 20\% & 91.78(0.28) & &  10\% & 91.41(0.23) \\
DPF  & 20\% &  92.17(0.21) & &  10\% &  92.42(0.18) \\
  \midrule
BNN\_average & {9.88\%(0.02\%)} & 91.65(0.08) & & 4.77\%(0.08\%) & { 91.30(0.16)} \\ 
BNN\_anneal  & {9.95\%(0.03\%)} & 91.28(0.11) & & 4.88\%(0.02\%) & {91.17(0.08)}  \\ 
BNN\_BIC &  9.55\%(0.03\%) & {\ 91.27(0.05)} && 
4.78\%(0.01\%) & {91.21(0.01)}   \\ 
SM&  10\% & 89.76(0.40) & & 5\% & 88.68(0.22)   \\
DSR&  10\% & 87.88(0.04) & & 5\% & 84.12(0.32)  \\
DPF&  10\% &  90.88(0.07) & & 5\% &  90.94(0.35)  \\
\bottomrule
\end{tabular}
\end{center}
\end{table}

\begin{table*}[t]
\caption{ResNet network pruning results for CIFAR-10 data, which were calculated by averaging over 3 independent runs with the standard deviation reported in the parentheses.}
\label{CIFAR_calibration}
\begin{center}
\adjustbox{width=1.0\textwidth}{
\begin{tabular}{ccccccc}
 \toprule
Method & Model & Pruning Ratio & NLL & JS-Distance & ECE \\
  \midrule
DNN\_dense & ResNet20 & 100\% & 0.2276(0.0021) & 7.9118(0.9316) & 0.02627(0.0005) \\ \midrule
BNN\_average & ResNet20 & {9.88\%(0.02\%)} & 0.2528(0.0029) & 9.9641(0.3069) & 0.0113(0.0010)\\ 
BNN\_anneal  & ResNet20 & {9.95\%(0.03\%)}  & 0.2618(0.0037) & 10.1251(0.1797) & 0.0175(0.0011)  \\ 
DPF  & ResNet20 & 10\%  & 0.2833(0.0004)& 7.5712(0.4466)  & 0.0294(0.0009)\\
  \midrule
BNN\_average & ResNet20 & 19.85\%(0.18\%) & 0.2323(0.0033)& 7.7007(0.5374) & 0.0173(0.0014) \\ 
BNN\_anneal  & ResNet20 & 19.80\%(0.01\%)  & 0.2441(0.0042) & 6.4435(0.2029) & 0.0233(0.0020)\\ 
DPF & ResNet20 & 20\% &  0.2874(0.0029) & 7.7329(0.1400) & 0.0391(0.0001)\\
  \midrule \midrule
DNN\_dense & ResNet32 & 100\% & 0.2042(0.0017) & 6.7699(0.5253)  & 0.02613(0.00029)\\ \midrule
BNN\_average & ResNet32 & 9.99\%(0.08\%) & 0.2116(0.0012) & 9.4549(0.5456) & 0.0132(0.0001)  \\ 
BNN\_anneal  & ResNet32 & 9.97\%(0.03\%) & 0.2218(0.0013) & 8.5447(0.1393) & 0.0192(0.0009) \\ 
DPF & ResNet32 & 10\%  & 0.2677(0.0041)& 7.8693(0.1840) & 0.0364(0.0015)\\
  \midrule
BNN\_average & ResNet32 & 4.77\%(0.08\%) & 0.2587(0.0022) & 7.0117(0.2222) & 0.0100(0.0002) \\ 
BNN\_anneal  & ResNet32 & 4.88\%(0.02\%)  & 0.2676(0.0014)  & 6.8440(0.4850)  & 0.0149(0.0006) \\ 
DPF & ResNet32 & 5\%  &  0.2921(0.0067)& 6.3990(0.8384)  & 0.0276(0.0019)\\
 \bottomrule
\end{tabular}
}
\end{center}
\vspace{-0.2in}
\end{table*}

\section{Conclusion} 

This work, together with \cite{SunSLiang2021}, has built a  solid theoretical foundation for sparse deep learning, which has successfully tamed the 
sparse 
deep neural network into the framework of statistical modeling. As implied by Lemma \ref{2normal}, Lemma \ref{Selectlem}, Theorem \ref{bvm}, and Theorem \ref{predictionbvm}, the sparse DNN can be simply viewed as a nonlinear statistical model which, like a traditional statistical model, possesses many nice properties such as posterior consistency, variable selection consistency, and asymptotic normality.  
We have shown how the prediction uncertainty of the sparse DNN can be quantified based on the asymptotic normality theory, and 
provided algorithms for training sparse DNNs with theoretical guarantees for its convergence to the global optimum. The latter ensures the validity of the down-stream statistical inference.

\section*{Acknowledgment}
Funding in direct support of this work: NSF grant DMS-2015498, NIH grants R01-GM117597 and R01-GM126089, and Liang's startup fund at Purdue University.  

\bibliographystyle{plainnat}
\bibliography{reference}

\begin{thebibliography}{36}
\providecommand{\natexlab}[1]{#1}
\providecommand{\url}[1]{\texttt{#1}}
\expandafter\ifx\csname urlstyle\endcsname\relax
  \providecommand{\doi}[1]{doi: #1}\else
  \providecommand{\doi}{doi: \begingroup \urlstyle{rm}\Url}\fi

\bibitem[Allen-Zhu et~al.(2019)Allen-Zhu, Li, and Song]{AllenZhu2019ACT}
Zeyuan Allen-Zhu, Yuanzhi Li, and Zhao Song.
\newblock A convergence theory for deep learning via over-parameterization.
\newblock In \emph{ICML}, 2019.

\bibitem[Bleich et~al.(2014)Bleich, Kapelner, George, and
  Jensen]{bleich2014variable}
Justin Bleich, Adam Kapelner, Edward~I George, and Shane~T Jensen.
\newblock Variable selection for bart: an application to gene regulation.
\newblock \emph{The Annals of Applied Statistics}, pages 1750--1781, 2014.

\bibitem[B{\"o}lcskei et~al.(2019)B{\"o}lcskei, Grohs, Kutyniok, and
  Petersen]{Bolcskei2019}
Helmut B{\"o}lcskei, Philipp Grohs, Gitta Kutyniok, and Philipp Petersen.
\newblock Optimal approximation with sparsely connected deep neural networks.
\newblock \emph{CoRR}, abs/1705.01714, 2019.

\bibitem[Castillo and Rousseau(2015)]{castillo2015supplement}
Isma{\"e}l Castillo and Judith Rousseau.
\newblock Supplement to ``a bernstein--von mises theorem for smooth functionals
  in semiparametric models''.
\newblock \emph{Annals of Statistics}, 43\penalty0 (6):\penalty0 2353--2383,
  2015.

\bibitem[Castillo et~al.(2015)Castillo, Rousseau,
  et~al.]{castillo2015bernstein}
Isma{\"e}l Castillo, Judith Rousseau, et~al.
\newblock A bernstein--von mises theorem for smooth functionals in
  semiparametric models.
\newblock \emph{The Annals of Statistics}, 43\penalty0 (6):\penalty0
  2353--2383, 2015.

\bibitem[Chen et~al.(2015)Chen, Ding, and Carin]{chen2015convergence}
Changyou Chen, Nan Ding, and Lawrence Carin.
\newblock On the convergence of stochastic gradient mcmc algorithms with
  high-order integrators.
\newblock In \emph{Proceedings of the 28th International Conference on Neural
  Information Processing Systems-Volume 2}, pages 2278--2286, 2015.

\bibitem[Chen et~al.(2014)Chen, Fox, and Guestrin]{SGHMC2014}
Tianqi Chen, Emily Fox, and Carlos Guestrin.
\newblock Stochastic gradient hamiltonian monte carlo.
\newblock In \emph{International conference on machine learning}, pages
  1683--1691, 2014.

\bibitem[Dettmers and Zettlemoyer(2019)]{dettmers2019sparse}
Tim Dettmers and Luke Zettlemoyer.
\newblock Sparse networks from scratch: Faster training without losing
  performance.
\newblock \emph{arXiv preprint arXiv:1907.04840}, 2019.

\bibitem[Du et~al.(2019)Du, Lee, Li, Wang, and Zhai]{Du2019GradientDF}
Simon~S. Du, Jason~D. Lee, Haochuan Li, Liwei Wang, and Xiyu Zhai.
\newblock Gradient descent finds global minima of deep neural networks.
\newblock In \emph{ICML}, 2019.

\bibitem[Fan and Lv(2008)]{fan2008sure}
Jianqing Fan and Jinchi Lv.
\newblock Sure independence screening for ultrahigh dimensional feature space.
\newblock \emph{Journal of the Royal Statistical Society: Series B (Statistical
  Methodology)}, 70\penalty0 (5):\penalty0 849--911, 2008.

\bibitem[Fefferman(1994)]{fefferman1994reconstructing}
Charles Fefferman.
\newblock Reconstructing a neural net from its output.
\newblock \emph{Revista Matem{\'a}tica Iberoamericana}, 10\penalty0
  (3):\penalty0 507--555, 1994.

\bibitem[Feng and Simon(2017)]{feng2017sparse}
Jean Feng and Noah Simon.
\newblock Sparse-input neural networks for high-dimensional nonparametric
  regression and classification.
\newblock \emph{arXiv preprint arXiv:1711.07592}, 2017.

\bibitem[Gal and Ghahramani(2016)]{MCdropout2016}
Yarin Gal and Zoubin Ghahramani.
\newblock Dropout as a bayesian approximation: Representing model uncertainty
  in deep learning.
\newblock In \emph{Proceedings of the 33rd International Conference on
  International Conference on Machine Learning - Volume 48}, ICML'16, page
  1050–1059. JMLR.org, 2016.

\bibitem[{Gori} and {Tesi}(1992)]{GoriTesi1992}
M.~{Gori} and A.~{Tesi}.
\newblock On the problem of local minima in backpropagation.
\newblock \emph{IEEE Transactions on Pattern Analysis and Machine
  Intelligence}, 14\penalty0 (1):\penalty0 76--86, 1992.

\bibitem[Guo et~al.(2017)Guo, Pleiss, Sun, and Weinberger]{CalibrationDNN2017}
Chuan Guo, Geoff Pleiss, Yu~Sun, and Kilian~Q. Weinberger.
\newblock On calibration of modern neural networks.
\newblock In \emph{Proceedings of the 34th International Conference on Machine
  Learning - Volume 70}, ICML'17, page 1321–1330. JMLR.org, 2017.

\bibitem[He et~al.(2016)He, Zhang, Ren, and Sun]{he2016deep}
Kaiming He, Xiangyu Zhang, Shaoqing Ren, and Jian Sun.
\newblock Deep residual learning for image recognition.
\newblock In \emph{Proceedings of the IEEE conference on computer vision and
  pattern recognition}, pages 770--778, 2016.

\bibitem[Kingma and Ba(2015)]{adam2015}
D.P. Kingma and J.L. Ba.
\newblock Adam: a method for stochastic optimization.
\newblock In \emph{International Conference on Learning Representations}, 2015.

\bibitem[Krizhevsky et~al.(2009)Krizhevsky, Hinton,
  et~al.]{krizhevsky2009learning}
Alex Krizhevsky, Geoffrey Hinton, et~al.
\newblock Learning multiple layers of features from tiny images.
\newblock Technical report, Citeseer, 2009.

\bibitem[Lakshminarayanan et~al.(2017)Lakshminarayanan, Pritzel, and
  Blundell]{Deepensemble2017}
Balaji Lakshminarayanan, Alexander Pritzel, and Charles Blundell.
\newblock Simple and scalable predictive uncertainty estimation using deep
  ensembles.
\newblock In \emph{Proceedings of the 31st International Conference on Neural
  Information Processing Systems}, NIPS'17, page 6405–6416, Red Hook, NY,
  USA, 2017. Curran Associates Inc.
\newblock ISBN 9781510860964.

\bibitem[Liang et~al.(2013)Liang, Song, and Yu]{LiangSY2013}
F.~Liang, Q.~Song, and K.~Yu.
\newblock Bayesian subset modeling for high dimensional generalized linear
  models.
\newblock \emph{\JASA}, 108:\penalty0 589--606, 2013.

\bibitem[Liang et~al.(2018)Liang, Li, and Zhou]{liang2018bayesian}
F.~Liang, Q.~Li, and L.~Zhou.
\newblock Bayesian neural networks for selection of drug sensitive genes.
\newblock \emph{Journal of the American Statistical Association}, 113\penalty0
  (523):\penalty0 955--972, 2018.

\bibitem[Lin et~al.(2020)Lin, Stich, Barba, Dmitriev, and
  Jaggi]{lin2020dynamic}
Tao Lin, Sebastian~U. Stich, Luis Barba, Daniil Dmitriev, and Martin Jaggi.
\newblock Dynamic model pruning with feedback.
\newblock In \emph{International Conference on Learning Representations}, 2020.
\newblock URL \url{https://openreview.net/forum?id=SJem8lSFwB}.

\bibitem[Ma et~al.(2015)Ma, Chen, and Fox]{ma2015complete}
Yi-An Ma, Tianqi Chen, and Emily Fox.
\newblock A complete recipe for stochastic gradient mcmc.
\newblock In \emph{Advances in Neural Information Processing Systems}, pages
  2917--2925, 2015.

\bibitem[Maddox et~al.(2019)Maddox, Izmailov, Garipov, Vetrov, and
  Wilson]{maddox2019simple}
Wesley~J Maddox, Pavel Izmailov, Timur Garipov, Dmitry~P Vetrov, and
  Andrew~Gordon Wilson.
\newblock A simple baseline for bayesian uncertainty in deep learning.
\newblock In \emph{Advances in Neural Information Processing Systems}, pages
  13153--13164, 2019.

\bibitem[Mostafa and Wang(2019)]{mostafa2019parameter}
Hesham Mostafa and Xin Wang.
\newblock Parameter efficient training of deep convolutional neural networks by
  dynamic sparse reparameterization.
\newblock In \emph{International Conference on Machine Learning}, pages
  4646--4655, 2019.

\bibitem[Nemeth and Fearnhead(2019)]{nemeth2019stochastic}
Christopher Nemeth and Paul Fearnhead.
\newblock Stochastic gradient markov chain monte carlo.
\newblock \emph{arXiv preprint arXiv:1907.06986}, 2019.

\bibitem[Nguyen and Hein(2017)]{Nguyen2017TheLS}
Quynh Nguyen and Matthias Hein.
\newblock The loss surface of deep and wide neural networks.
\newblock In \emph{ICML}, 2017.

\bibitem[Polson and Ro\v{c}kov\'{a}(2018)]{polson2018posterior}
Nicholas~G. Polson and Veronika Ro\v{c}kov\'{a}.
\newblock Posterior concentration for sparse deep learning.
\newblock In \emph{Proceedings of the 32nd International Conference on Neural
  Information Processing Systems}, NIPS'18, page 938–949, Red Hook, NY, USA,
  2018. Curran Associates Inc.

\bibitem[Portnoy(1988)]{Portnoy1988}
S.~Portnoy.
\newblock Asymptotic behavior of likelihood methods for exponential families
  when the number of parameters tend to infinity.
\newblock \emph{\ANNALS}, 16\penalty0 (1):\penalty0 356--366, 1988.

\bibitem[Schmidt-Hieber(2017)]{Schmidt-Hieber2017Nonparametric}
Johannes Schmidt-Hieber.
\newblock Nonparametric regression using deep neural networks with relu
  activation function.
\newblock \emph{arXiv:1708.06633}, 2017.

\bibitem[Sun et~al.(2021)Sun, Song, and Liang]{SunSLiang2021}
Y.~Sun, Q.~Song, and F.~Liang.
\newblock Consistent sparse deep learning: Theory and computation.
\newblock \emph{Journal of the American Statistical Association}, page in
  press, 2021.

\bibitem[Tibshirani(1996)]{tibshirani1996regression}
Robert Tibshirani.
\newblock Regression shrinkage and selection via the lasso.
\newblock \emph{Journal of the Royal Statistical Society. Series B
  (Methodological)}, pages 267--288, 1996.

\bibitem[Wang and Rockov{\'a}(2020)]{Wang2020UncertaintyQF}
Yuexi Wang and V.~Rockov{\'a}.
\newblock Uncertainty quantification for sparse deep learning.
\newblock In \emph{AISTATS}, 2020.

\bibitem[Welling and Teh(2011)]{welling2011bayesian}
Max Welling and Yee~W Teh.
\newblock Bayesian learning via stochastic gradient langevin dynamics.
\newblock In \emph{Proceedings of the 28th international conference on machine
  learning (ICML-11)}, pages 681--688, 2011.

\bibitem[Zou and Gu(2019)]{ZouGu2019}
Difan Zou and Quanquan Gu.
\newblock An improved analysis of training over-parameterized deep neural
  networks.
\newblock In \emph{NuerIPS}, 2019.

\bibitem[Zou et~al.(2020)Zou, Cao, Zhou, and Gu]{Zou2020GradientDO}
Difan Zou, Yuan Cao, Dongruo Zhou, and Quanquan Gu.
\newblock Gradient descent optimizes over-parameterized deep relu networks.
\newblock \emph{Machine Learning}, 109:\penalty0 467 -- 492, 2020.

\end{thebibliography}

\newpage
\appendix

\section{Supplementary material of "Sparse Deep Learning: A New Framework Immune to Local Traps and Miscalibration"}

\subsection{Proof of Theorem 2.1}

\proof We first define the equivalent class of neural network parameters. Given a parameter vector $\bbeta$ and the corresponding structure parameter vector $\bgamma$, its equivalent class is given by 
\[
Q_E(\bbeta, \bgamma) = \{(\tilde{\bbeta}, \tilde{\bgamma}):
\nu_g(\tilde{\bbeta},\tilde{\bgamma})=(\bbeta,\bgamma),
\mu(\tilde{\bbeta}, \tilde{\bgamma}, \bx) = \mu(\bbeta,\bgamma, \bx), \forall \bx \},
\]
where $\nu_g(\cdot)$ denotes a generic mapping that contains only the transformations of node permutation and weight sign flipping. Specifically, we define
 \[
 Q_E^* = Q_E(\bbeta^*, \bgamma^*),
 \]
which represents the equivalent class of {\it true DNN model}. 

Let $B_{\delta_n}(\bbeta^*) = \{\bbeta: |\bbeta_i-\bbeta^*_i| < \delta_n, \forall i\in \bgamma^*, |\bbeta_i-\bbeta^*_i| < 2\sigma_{0,n}\log(\frac{\sigma_{1,n}}{\lambda_n\sigma_{0,n}}),\forall i \notin \bgamma^* \}$.
By assumption C.1, $\bbeta^*$ is generic (i.e. $Q_E(\bbeta^*)$ contains only reparameterizations of weight sign-flipping or node permutations as defined in \cite{feng2017sparse} and  \cite{fefferman1994reconstructing}) and $\min_{i\in \bgamma^*}|\bbeta^*_i| - \delta_n > (C-1)\delta_n >\delta_n$,
then for any $\bbeta^{*(1)}, \bbeta^{*(2)} \in Q_E^*$,  $B_{\delta_n}(\bbeta^{*(1)}) \cap B_{\delta_n}(\bbeta^{*(2)}) = \emptyset$, and thus  $\{\bbeta: \tilde{\nu}(\bbeta) \in B_{\delta_n}(\bbeta^*)\} = \cup_{\bbeta\in Q_E^*} B_{\delta_n}(\bbeta)$.
In what follows, we 
 will first show $\pi(\cup_{\bbeta\in Q_E^*} B_{\delta_n}(\bbeta) \mid D_n) \rightarrow 1$ as 
 $n\to \infty$, which means the most posterior mass falls in the neighbourhood of true parameter.
 
  \underline{Remark on the notation}: $\tilde{\nu}(\cdot)$ is similar to  $\nu(\cdot)$ defined in Section 2.1 of the main text. They both map the set $Q_E(\bbeta,\bgamma)$ 
   to a unique network. The difference between them is that $\|\nu(\bbeta)-\bbeta^*\|_{\infty}$ may be arbitrary, but  $\|\tilde{\nu}(\bbeta)-\bbeta^*\|_{\infty}$ is minimized.  In other words, $\nu(\bbeta,\bgamma)$ is to find an arbitrary network in   $Q_E(\bbeta,\bgamma)$ as the representative of the equivalent class, 
   while $\tilde{\nu}(\bbeta,\bgamma)$ is to find a  representative in  $Q_E(\bbeta,\bgamma)$ such that the distance between 
  $\bbeta^*$ and the representative is minimized. In what follows, we will use $\tilde{\nu}(\bbeta)$ 
   and $\tilde{\nu}(\bgamma)$ to denote the connection weight and network structure of $\tilde{\nu}(\bbeta,\bgamma)$, respectively. 
   With a slight abuse of notation, we will use 
   $\tilde{\nu}(\bbeta)_i$ to denote the $i$th component of $\tilde{\nu}(\bbeta)$, and use 
   $\tilde{\nu}(\bgamma)_i$ to denote the $i$th component of $\tilde{\nu}(\bgamma)$. 
  
Recall that the marginal posterior inclusion probability is given by 
\[
q_i =\int \sum_{\bgamma} e_{i|\tilde{\nu}(\bbeta,\bgamma)} \pi(\bgamma|\bbeta,D_n)\pi(\bbeta|D_n) d\bbeta
=\int \pi(\tilde{\nu}(\bgamma)_{i}=1|\bbeta)\pi(\bbeta|D_n)d \bbeta.
\]
For the mixture Gaussian prior, 
\[
\pi(\bgamma_i = 1 | \bbeta) = \frac{1}{1 + \frac{\sigma_{1,n} (1-\lambda_n)}{\sigma_{0,n}\lambda_n} e^{-(\frac{1}{2\sigma_{0,n}^2} - \frac{1}{2\sigma_{1,n}^2})\bbeta_i^2 }},
\]
which increases with respect to $|\bbeta_i|$. In particular,  if  $|\bbeta_i| > 2\sigma_{0,n}\log(\frac{\sigma_{1,n}}{\lambda_n\sigma_{0,n}})$, then $\pi(\bgamma_i = 1 | \bbeta) > \frac{1}{2}$.

For the mixture Gaussian prior,
\begin{equation}
\nonumber
\begin{split}
&\pi(\bbeta \notin \cup_{\bbeta\in Q_E^*} B_{\delta_n}(\bbeta) \mid D_n)\\
\leq & \pi(\exists i \notin \bgamma^{*}, |\tilde{\nu}(\bbeta)_i| > 2\sigma_{0,n}\log(\frac{\sigma_{1,n}}{\lambda_n\sigma_{0,n}}) \mid D_n) + \pi(\exists i \in \bgamma^{*}, |\tilde{\nu}(\bbeta)_i - \bbeta^{*}_i| > \delta_n \mid D_n).
\end{split}
\end{equation}
For the first term, note that for a given $i\notin \bgamma^{*}$,
\begin{equation}
\nonumber
\begin{split}
 \pi(|\tilde{\nu}(\bbeta)_i| > 2\sigma_{0,n}\log(\frac{\sigma_{1,n}}{\lambda_n\sigma_{0,n}}) \mid D_n) 
 \leq & \pi(\pi(\tilde{\nu}(\bgamma)_{i}=1|\bbeta) > \frac{1}{2} \mid D_n) \\ 
\leq & 2\int  \pi(\tilde{\nu}(\bgamma)_{i}=1|\bbeta)\pi(\bbeta|D_n)d \bbeta \\
\leq & 2 \rho(\epsilon_n) + 2 \pi(d(p_\bbeta,p_{\mu^*}) \geq \epsilon_n \mid D_n)
\rightarrow 0. 
\end{split}
\end{equation}
Then we have 
\begin{equation}
\nonumber
\begin{split}
 \pi(\exists i \notin \bgamma^{*}, |\tilde{\nu}(\bbeta)_i| > 2\sigma_{0,n}\log(\frac{\sigma_{1,n}}{\lambda_n\sigma_{0,n}}) \mid D_n) 
 = &  \pi(\max_{i \notin \bgamma^{*}} |\tilde{\nu}(\bbeta)_i| > 2\sigma_{0,n}\log(\frac{\sigma_{1,n}}{\lambda_n\sigma_{0,n}}) \mid D_n)\\
\leq & \pi(\max_{i \notin \bgamma^{*}}\pi(\tilde{\nu}(\bgamma)_{i}=1|\bbeta) > \frac{1}{2} \mid D_n) \\ 
\leq & \sum_{i\notin \bgamma^{*}} \pi(\pi(\tilde{\nu}(\bgamma)_{i}=1|\bbeta) > \frac{1}{2} \mid D_n) \\
\leq & 2 K_n\rho(\epsilon_n) + 2 K_n \pi(d(p_\bbeta,p_{\mu^*}) \geq \epsilon_n \mid D_n)
\rightarrow 0. 
\end{split}
\end{equation}
For the second term, by condition C.1 and
\textcolor{black}{Lemma 2.1},
\begin{equation}
\nonumber
\begin{split}
& \pi(\exists i \in \bgamma^{*}, |\tilde{\nu}(\bbeta)_i - \bbeta^{*}_i| > \delta_n \mid D_n)  
=  \pi(\max_{i \in \bgamma^{*}} |\tilde{\nu}(\bbeta)_i - \bbeta^{*}_i| > \delta_n \mid D_n) \\
= & \pi(\max_{i \in \bgamma^{*}} |\tilde{\nu}(\bbeta)_i - \bbeta^{*}_i| > \delta_n,  d(p_\bbeta,p_{\mu^*}) \leq \epsilon_n \mid D_n) \\
 & + \pi(\max_{i \in \bgamma^{*}} |\tilde{\nu}(\bbeta)_i - \bbeta^{*}_i| > \delta_n,  d(p_\bbeta,p_{\mu^*}) \geq \epsilon_n \mid D_n) \\
\leq & \pi(A(\epsilon_n, \delta_n) \mid D_n) + \pi(d(p_\bbeta,p_{\mu^*}) \geq \epsilon_n \mid D_n) \rightarrow 0.
\end{split}
\end{equation}
Summarizing the above two terms, we have  $\pi(\cup_{\bbeta\in Q_E^*} B_{\delta_n}(\bbeta) \mid D_n) \rightarrow 1$.

Let $Q_n = |Q_E^*|$ be the number of equivalent {\it true DNN model}. By the generic assumption of $\bbeta^*$, 
for any $\bbeta^{*(1)}, \bbeta^{*(2)} \in Q_E^*$,  $B_{\delta_n}(\bbeta^{*(1)}) \cap B_{\delta_n}(\bbeta^{*(2)}) = \emptyset$. Then in $B_{\delta_n}(\bbeta^{*})$, the posterior density of $\tilde{\nu}(\bbeta)$ is $Q_n$ times the posterior density of $\bbeta$. 
Then for a function $f(\cdot)$ of $\tilde{\nu}(\bbeta)$, by changing variable,
\[
\int_{\tilde{\nu}(\bbeta)\in B_{\delta_n}(\bbeta^*)} f(\tilde{\nu}(\bbeta)) \pi(\tilde{\nu}(\bbeta)|D_n) d\tilde{\nu}(\bbeta) = Q_n \int_{B_{\delta_n}(\bbeta^*)} f(\bbeta)\pi(\bbeta|D_n) d\bbeta.
\]
Thus, we only need to consider the integration on $B_{\delta_n}(\bbeta^*)$. Define 
 \[
\hat{\bbeta}_i = \begin{cases}
 \bbeta^*_i - \sum_{j\in \bgamma^*}h^{i,j}(\bbeta^*) h_j(\bbeta^*), &  \forall i \in \bgamma^*,  \\
  0, &  \forall i \not\in \bgamma^*. \\
 \end{cases}
 \]
We will first prove that for any real vector $\bt$,
\begin{equation}
\begin{split}
E(e^{\sqrt{n} \bt^{T}(\tilde{\nu}(\bbeta)-\hat{\bbeta})} \mid D_n,  B_{\delta_n}(\bbeta^*)) 
:= & \frac{\int_{ B_{\delta_n}(\bbeta^*)} e^{\sqrt{n} \bt^{T} (\tilde{\nu}(\bbeta)-\hat{\bbeta})} \pi(\tilde{\nu}(\bbeta) | D_n) d\tilde{\nu}(\bbeta) }{\int_{ B_{\delta_n}(\bbeta^*)} \pi(\tilde{\nu}(\bbeta) | D_n) d\tilde{\nu}(\bbeta)}\\
= &\frac{\int_{ B_{\delta_n}(\bbeta^*)} e^{\sqrt{n}\bt^{T} (\bbeta-\hat{\bbeta})} e^{nl_n(\bbeta)}\pi(\bbeta) d\bbeta }{\int_{ B_{\delta_n}(\bbeta^*)} e^{nl_n(\bbeta)}\pi(\bbeta) d\bbeta}\\
=&e^{\frac{1}{2}\bt^T\bV\bt + o_{P^*}(1)}.
\end{split}
\end{equation}

For any $\bbeta \in B_{\delta_n}(\bbeta^*)$, we have
\begin{equation}
\nonumber
\begin{split}
 |\sqrt{n}(\bt^{T} (\bbeta-\bbeta_{\bgamma^*}))|
\leq  \sqrt{n}K_n ||\bt||_{\infty} 2\sigma_{0,n}\log(\frac{\sigma_{1,n}}{\lambda_n\sigma_{0,n}}) 
= o(1),
\end{split}
\end{equation}
\begin{equation}
\nonumber
\begin{split}
|n(l_n(\bbeta)-l_n(\bbeta_{\bgamma^*}))| = |n\sum_{i\notin \bgamma^*}\bbeta_i(h_i(\tilde{\bbeta}))| 
\leq  nK_n M 2\sigma_{0,n}\log(\frac{\sigma_{1,n}}{\lambda_n\sigma_{0,n}}) 
= o(1).
\end{split}
\end{equation}
Then, we have 
\begin{equation} \label{reveq1}
\begin{split}
 \sqrt{n} \bt^{T}(\bbeta - \hat{\bbeta}) 
= & \sqrt{n} \bt^{T}(\bbeta - \bbeta_{\bgamma^*} + \bbeta_{\bgamma^*} - \bbeta^*)  + \sqrt{n}\sum_{i,j\in \bgamma^*}h^{i,j}(\bbeta^*)\bt_j h_i(\bbeta^*))\\
= & o(1) + \sqrt{n}\sum_{i\in \bgamma^*}(\bbeta_i - \bbeta^*_i)\bt_i + \sqrt{n}\sum_{i,j\in \bgamma^*}h^{i,j}(\bbeta^*)\bt_j h_i(\bbeta^*),
\end{split}
\end{equation}
\begin{equation} \label{reveq2}
\begin{split}
 nl_n(\bbeta) - nl_n(\bbeta^*) 
= & n(l_n(\bbeta) - l_n(\bbeta_{\bgamma^*}) + l_n(\bbeta_{\bgamma^*})- nl_n(\bbeta^*)) \\
 =& o(1) + n\sum_{i \in \bgamma^*} (\bbeta_i - \bbeta^*_i)h_i(\bbeta^*) + \frac{n}{2}\sum_{i,j\in \bgamma^*} h_{i,j}(\bbeta^*)(\bbeta_i - \bbeta_i^*)(\bbeta_j - \bbeta^*_j)\\
 &+ \frac{n}{6}\sum_{i,j,k\in \bgamma^*} h_{i,j,k}(\check{\bbeta})(\bbeta_i - \bbeta_i^*)(\bbeta_j - \bbeta^*_j) (\bbeta_k - \bbeta^*_k), 
\end{split}
\end{equation}
where $ \check{\bbeta}$ is a point between $\bbeta_{\bgamma^*}$ and $\bbeta^*$. Note that for $\bbeta \in B_{\delta_n}(\bbeta^*)$, $|\bbeta_i - \bbeta^*_i| \leq \delta_n \lesssim \frac{1}{\sqrt[3]{n}r_n}$, we have $\frac{n}{6}\sum_{i,j,k\in \bgamma^*} h_{i,j,k}(\check{\bbeta})(\bbeta_i - \bbeta_i^*)(\bbeta_j - \bbeta^*_j) (\bbeta_k - \bbeta^*_k) = o(1) $.

Let $\bbeta^{(t)}$ be network parameters satisfying $\bbeta^{(t)}_i = \bbeta_i + \frac{1}{\sqrt{n}} \sum_{j\in \bgamma^*} h^{i,j}(\bbeta^*)\bt_j, \forall i \in \bgamma^*$ and $\bbeta^{(t)}_i = \bbeta_i, \forall i \notin \bgamma^*$. Note that $\frac{1}{\sqrt{n}} \sum_{j\in \bgamma^*} h^{i,j}(\bbeta^*)\bt_j \leq \frac{r_n ||\bt||_{\infty} M}{\sqrt{n}} \lesssim \delta_n$, for large enough $n$, $|\bbeta^{(t)}_i| < 2\delta_n$ $\forall i\in \bgamma^*$. Thus, we have 
\begin{equation} \label{reveq3}
\begin{split}
 nl_n(\bbeta^{(t)}) - nl_n(\bbeta^*) 
= & n(l_n(\bbeta^{(t)}) - l_n(\bbeta^{(t)}_{\bgamma^*}) + l_n(\bbeta^{(t)}_{\bgamma^*})- nl_n(\bbeta^*)) \\
 =& o(1) + n\sum_{i \in \bgamma^*} (\bbeta^{(t)}_i - \bbeta^*_i)h_i(\bbeta^*) + \frac{n}{2}\sum_{i,j\in \bgamma^*} h_{i,j}(\bbeta^*)(\bbeta^{(t)}_i - \bbeta_i^*)(\bbeta^{(t)}_j - \bbeta^*_j)\\
 =& o(1) + n\sum_{i \in \bgamma^*} (\bbeta_i - \bbeta^*_i)h_i(\bbeta^*) + \frac{n}{2}\sum_{i,j\in \bgamma^*} h_{i,j}(\bbeta^*)(\bbeta_i - \bbeta_i^*)(\bbeta_j - \bbeta^*_j) \\
 &+  \sqrt{n}\sum_{i,j\in \bgamma^*}h^{i,j}(\bbeta^*)\bt_j h_i(\bbeta^*) + \sqrt{n}\sum_{i\in \bgamma^*}(\bbeta_i - \bbeta^*_i)\bt_i + \frac{1}{2}\sum_{i,j\in \bgamma^*}h^{i,j}(\bbeta^*)\bt_i \bt_j\\
 =& o(1) +  \sqrt{n}\bt^{T}(\bbeta - \hat{\bbeta}) + nl_n(\bbeta) - nl_n(\bbeta^*) + \frac{1}{2}\sum_{i,j\in \bgamma^*}h^{i,j}(\bbeta^*) \bt_i \bt_j,
\end{split}
\end{equation}
where the last equality is derived by replacing appropriate terms by $\sqrt{n}\bt^{T}(\bbeta - \hat{\bbeta})$ and $nl_n(\bbeta) - nl_n(\bbeta^*)$ based on (\ref{reveq1}) and (\ref{reveq2}), respectively; and the third equality is derived based on the following calculation: 
\begin{equation}
\begin{split}
& \frac{n}{2}\sum_{i,j\in \bgamma^*} h_{i,j}(\bbeta^*)(\bbeta^{(t)}_i - \bbeta_i^*)(\bbeta^{(t)}_j - \bbeta^*_j) \\
= & \frac{n}{2}\sum_{i,j\in \bgamma^*} h_{i,j}(\bbeta^*) (\bbeta_i - \bbeta_i^* + \frac{1}{\sqrt{n}} \sum_{k\in \bgamma^*} h^{i,k}(\bbeta^*)\bt_k)(\bbeta_j - \bbeta_j^* + \frac{1}{\sqrt{n}} \sum_{k\in \bgamma^*} h^{j,k}(\bbeta^*)\bt_k) \\
= & \frac{n}{2}\sum_{i,j\in \bgamma^*} h_{i,j}(\bbeta^*)(\bbeta_i - \bbeta_i^*)(\bbeta_j - \bbeta_j^*) + 2 \times \frac{n}{2}\sum_{i,j\in \bgamma^*} h_{i,j}(\bbeta^*)\frac{1}{\sqrt{n}} \sum_{k\in \bgamma^*} h^{i,k}(\bbeta^*)\bt_k(\bbeta_j - \bbeta_j^*) \\
 & + \frac{n}{2}\sum_{i,j\in \bgamma^*} h_{i,j}(\bbeta^*)(\frac{1}{\sqrt{n}} \sum_{k\in \bgamma^*} h^{i,k}(\bbeta^*)\bt_k)(\frac{1}{\sqrt{n}} \sum_{k\in \bgamma^*} h^{j,k}(\bbeta^*)\bt_k) \\
 = & \frac{n}{2}\sum_{i,j\in \bgamma^*} h_{i,j}(\bbeta^*)(\bbeta_i - \bbeta_i^*)(\bbeta_j - \bbeta_j^*) + 
  \sqrt{n}\sum_{i\in \bgamma^*}(\bbeta_i - \bbeta^*_i)\bt_i + \frac{1}{2}\sum_{i,j\in \bgamma^*}h^{i,j}(\bbeta^*)\bt_i \bt_j, \\
\end{split}
\end{equation}
where the second and third terms  in the last equality are derived based on the relation $\sum_{i \in \bgamma^*}h_{i,j}(\bbeta^*) h^{i,k}(\bbeta^*) = \delta_{j,k}$, where $\delta_{j,k} = 1$ if $j = k$, $\delta_{j,k} = 0$ if $j \neq k$.  

By rearranging the terms in (\ref{reveq3}), we have  
\begin{equation*}
\begin{split}
& \int_{B_{\delta_n}(\bbeta^*)} \exp\{\sqrt{n} \bt^{T}(\bbeta-\hat{\bbeta})+ nl_n(\bbeta)\}\pi(\bbeta) d\bbeta \\
& = \exp\left\{-\frac{1}{2}\sum_{i,j\in \bgamma^*}h^{i,j}(\bbeta^*) \bt_i \bt_j + o(1)\right\} \int_{B_{\delta_n}(\bbeta^*)}  e^{nl_n(\bbeta^{(t)})}\pi(\bbeta) d\bbeta.
\end{split}
\end{equation*}
For $\bbeta \in B_{\delta_n}(\bbeta^*),i\in \bgamma^*$, by Assumption C.1, there exists a constant $C>2$ such that 
\[
\begin{split}
|\bbeta^{(t)}_i| & \geq |\bbeta_i| -  \frac{r_n ||\bt||_{\infty} M}{\sqrt{n}} \geq |\bbeta^*_i| - 2\delta_n \geq (C-2)\delta_n \gtrsim \frac{r_n}{\sqrt{n}} \\ 
& \gtrsim \sqrt{\left(\frac{1}{2\sigma_{0,n}^2} - \frac{1}{2\sigma_{1,n}^2} \right)^{-1} \log\left(\frac{r_n(1-\lambda_n)\sigma_{1,n}}{\sigma_{0,n} \lambda_n} \right)}.
\end{split}
\]
Then we have 
\[
  \frac{\sigma_{1,n} (1-\lambda_n)}{\sigma_{0,n}\lambda_n} e^{-(\frac{1}{2\sigma_{0,n}^2} - \frac{1}{2\sigma_{1,n}^2})(\bbeta^{(t)}_i)^2 } \lesssim \frac{1}{r_n}.
 \]
It is easy to see that the above formula also holds if we replace $\bbeta^{(t)}_i$ by $\bbeta_i$. Note that the mixture Gaussian prior of $\bbeta_i$ can be written as 
\[
\pi(\bbeta_i) = \frac{\lambda_n}{\sqrt{2\pi} \sigma_{1,n}}e^{-\frac{\bbeta_i^2}{2\sigma_{1,n}^2}}\left(1 + \frac{\sigma_{1,n} (1-\lambda_n)}{\sigma_{0,n}\lambda_n} e^{-(\frac{1}{2\sigma_{0,n}^2} - \frac{1}{2\sigma_{1,n}^2})\bbeta_i^2 } \right).
\]
Since $|\bbeta_i - \bbeta^{(t)}_i| \lesssim \delta_n \lesssim \frac{1}{\sqrt[3]{n} r_n}$, $|\bbeta_i + \bbeta^{(t)}_i| < 2E_n + 3\delta_n \lesssim E_n$,  and $\frac{1}{\sigma_{1,n}^2} \lesssim \frac{H_n\log(n) + \log(\bar{L})}{E_n^2}$, we have 
\[
\frac{r_n}{\sigma_{1,n}^2}(\bbeta_i - \bbeta^{(t)}_i)(\bbeta_i + \bbeta^{(t)}_i) = \frac{H_n\log(n)+\log(\bar{L})}{n^{C_1+1/3}}=o(1),
\]
by the condition $C_1>2/3$ and $H_n\log(n)+\log(\bar{L}) \prec n^{1-\epsilon}$. 
Thus, $\frac{\pi(\bbeta)}{\pi(\bbeta^{(t)})} = \prod_{i\in \bgamma^*} \frac{\pi(\bbeta_i)}{\pi(\bbeta^{(t)}_i)} = 1 + o(1)$, and 
\begin{equation}
\begin{split}
 \int_{B_{\delta_n}(\bbeta^*)}  e^{nl_n(\bbeta^{(t)})}\pi(\bbeta) d\bbeta 
= & (1+o(1)) \int_{\bbeta^{(t)}\in B_{\delta_n}(\bbeta^*)}  e^{nl_n(\bbeta^{(t)})}\pi(\bbeta^{(t)}) d\bbeta^{(t)} \\
= & (1+o(1)) C_N \pi(\bbeta^{(t)} \in B_{\delta_n}(\bbeta^*) \mid D_n),
\end{split}
\end{equation}
where $C_N$ is the normalizing constant of the posterior. Note that $||\bbeta^{(t)} - \bbeta||_{\infty} \lesssim \delta_n$, we have $\pi(\bbeta^{(t)} \in B_{\delta_n}(\bbeta^*) \mid D_n) \rightarrow  \pi(\bbeta \in B_{\delta_n}(\bbeta^*) \mid D_n)$. Moreover, since
$-\frac{1}{2}\sum_{i,j\in \bgamma^*}h^{i,j}(\bbeta^*) \bt_i \bt_j \rightarrow \frac{1}{2}\bt^{T}\bV \bt$, we have 
\[
E(e^{\sqrt{n} \bt^{T}(\tilde{\nu}(\bbeta)-\hat{\bbeta})} \mid D_n,  B_{\delta_n}(\bbeta^*))  = \frac{\int_{ B_{\delta_n}(\bbeta^*)} e^{\sqrt{n}\bt^{T} (\bbeta-\hat{\bbeta})} e^{nh_n(\bbeta)}\pi(\bbeta) d\bbeta }{\int_{ B_{\delta_n}(\bbeta^*)} e^{nh_n(\bbeta)}\pi(\bbeta) d\bbeta}=e^{\frac{\bt^{T}\bV \bt}{2} + o_{P^*}(1)}.
\]
Combining the above result with the fact that $\pi(\tilde{\nu}(\bbeta) \in B_{\delta_n}(\bbeta^*) \mid D_n) \rightarrow 1$, by section 1 of  \cite{castillo2015supplement}, we have 
\[
\pi[\sqrt n (\tilde{\nu}(\bbeta)-\hat{\bbeta}) \mid D_n] \rightsquigarrow N(0, \bV).
\]
We will then show that $\hat{\bbeta}$ will converge to $\bbeta^*$, then essentially we can replace $\hat{\bbeta}$ by $\bbeta^*$ in the above result. 
Let $\bTheta_{\bgamma^*} = \{\bbeta: \bbeta_i = 0, \forall i \notin \bgamma^*\}$ be the parameter space given the model $\bgamma^*$, and let $\hat{\bbeta}_{\bgamma^*}$ be the maximum likelihood estimator given the model $\bgamma^*$, i.e.
 \[
 \hat{\bbeta}_{\bgamma^*} = \arg\max_{\bbeta \in \bTheta_{\bgamma^*}} l_n(\bbeta).
 \]
Given condition C.3 and by Theorem 2.1 of \cite{Portnoy1988}, we have $||\hat{\bbeta}_{\bgamma^*} - \bbeta^* || = O(\sqrt{\frac{r_n}{n}}) = o(1)$.\\
Note that $h_i(\hat{\bbeta}_{\bgamma^*}) = 0$ as $\hat{\bbeta}_{\bgamma^*}$ is maximum likelihood estimator. Then for any $i\in \bgamma^*$, $|h_i(\bbeta^*)| = |h_i(\hat{\bbeta}_{\bgamma^*}) - h_i(\bbeta^*)| = |\sum_{j\in \bgamma^*} h_{ij}(\tilde{\bbeta})((\hat{\bbeta}_{\bgamma^*})_j - \bbeta^*_j)| \leq M||\hat{\bbeta}_{\bgamma^*} - \bbeta^* ||_1 = O(\sqrt{\frac{r_n}{n}})$.\\
Then for any $i,j \in \bgamma^*$, we have $\sum_{j\in \bgamma^*}h^{i,j}(\bbeta^*) h_j(\bbeta^*) = O(\sqrt{\frac{r_n^3}{n}}) = o(1)$. By the definition of $\hat{\bbeta}$, we have $\hat{\bbeta} - \bbeta^* = o(1)$. Therefore, we have 
\[
\pi[\sqrt n (\tilde{\nu}(\bbeta)-\bbeta^*) \mid D_n] \rightsquigarrow N(0, \bV).
\]

\subsection{Proof of Theorem 2.2}
\proof  The proof of Theorem 2.2 can be done using the same strategy as that used in proving Theorem 2.1. Here we provide a simpler proof using the result of Theorem 2.1. The notations we used in this proof are the same 
as in the proof of Theorem 2.1.
In the proof of Theorem 2.1, we have shown that $\pi(\tilde{\nu}(\bbeta) \in B_{\delta_n}(\bbeta^*) \mid D_n) \rightarrow 1$. 
Note that $\mu(\bbeta, \bx_0) = \mu(\tilde{\nu}(\bbeta), \bx_0)$. 
We only need to consider $\bbeta \in B_{\delta_n}(\bbeta^*)$. For $\bbeta \in B_{\delta_n}(\bbeta^*)$, we have
\begin{equation}
\nonumber
\begin{split}
& \sqrt{n}(\mu(\bbeta,\bx_0) - \mu(\bbeta^*,\bx_0)) \\
= & \sqrt{n}(\mu(\bbeta, \bx_0) - \mu(\bbeta_{\bgamma^*},\bx_0) + \mu(\tilde{\nu}(\bbeta_{\bgamma^*}),\bx_0) - \mu(\bbeta^*, \bx_0)).
\end{split}
\end{equation}
Since $\bbeta \in B_{\delta_n}(\bbeta^*)$, 
for $i\notin \bgamma^*$, $|\bbeta_i| < 2\sigma_{0,n}\log(\frac{\sigma_{1,n}}{\lambda_n\sigma_{0,n}})$; and for $i\in \bgamma^*$, $|\tilde{\nu}( \bbeta)_i - \bbeta^*_i| < \delta \lesssim \frac{1}{\sqrt[3]{n}r_n}$.  Therefore, 
\begin{equation}
\nonumber
\begin{split}
 |\sqrt{n}\mu(\bbeta,\bx_0)-\mu(\bbeta_{\bgamma^*}, \bx_0))|
= |\sqrt{n}\sum_{i\notin \bgamma^*}\bbeta_i(\mu_i(\tilde{\bbeta}, \bx_0))| 
\leq  \sqrt{n}K_n M 2\sigma_{0,n}\log(\frac{\sigma_{1,n}}{\lambda_n\sigma_{0,n}}) 
= o(1),
\end{split}
\end{equation}
where $\mu_i(\bbeta,\bx_0)$ denotes the first derivative of $\mu(\bbeta,\bx_0)$ with respect to the $i$th component of $\bbeta$, and $\tilde{\bbeta}$ denotes a point between $\bbeta$ and $\bbeta_{\bgamma^*}$. Further, 
\begin{equation}
\nonumber
\begin{split}
 &\mu(\tilde{\nu}(\bbeta_{\bgamma^*}),\bx_0) - \mu(\bbeta^*, \bx_0)\\
= & \sqrt{n}\sum_{i \in \bgamma^*}(\tilde{\nu}( \bbeta)_i - \bbeta^*_i)\mu_i(\bbeta^*, \bx_0) + \sqrt{n}\sum_{i \in \bgamma^*}\sum_{j \in \bgamma^*}(\tilde{\nu}( \bbeta)_i - \bbeta^*_i)\mu_{i,j}(\check{\bbeta},\bx_0) (\tilde{\nu}( \bbeta)_j - \bbeta^*_j) \\
= & \sqrt{n}\sum_{i \in \bgamma^*}((\tilde{\nu}( \bbeta)_i - \bbeta^*_i)\mu_i(\bbeta^*, \bx_0) + o(1),
\end{split}
\end{equation}
where
$\mu_{i,j}(\bbeta,\bx_0)$ denotes the second derivative of $\mu(\bbeta,\bx_0)$ with respect to the $i$th and $j$th components of $\bbeta$ and $\check{\bbeta}$ is a point between $\tilde{\nu}( \bbeta)$ and $\bbeta^*$.
Summarizing the above two equations, we have 
\begin{equation}
\nonumber
\begin{split}
\sqrt{n}\mu(\bbeta,\bx_0) - \mu(\bbeta^*,\bx_0)) = \sqrt{n}\sum_{i \in \bgamma^*}((\tilde{\nu}( \bbeta_i) - \bbeta^*_i)\mu_i(\bbeta^*, \bx_0) + o(1).
\end{split}
\end{equation}
By Theorem 2.1, $\pi[\sqrt n (\tilde{\nu}(\bbeta)-\bbeta^*) \mid D_n] \rightsquigarrow N(0, \bV)$, where $\bV=(v_{ij})$, and  $v_{i,j} = E(h^{i,j}(\bbeta^*))$ if $i,j \in \bgamma^*$ and $0$ otherwise. Then we have $\pi[\sqrt n (\mu(\bbeta, \bx_0)-\mu(\bbeta^*,\bx_0)) \mid D_n] \rightsquigarrow N(0, \Sigma)$,
where $\Sigma = \nabla_{\bgamma^*}\mu(\bbeta^*, \bx_0)^{T}H^{-1}\nabla_{\bgamma^*}\mu(\bbeta^*, \bx_0)$
 and  $H = E(-\nabla^2_{\bgamma^*} l_n(\bbeta^*))$.

\subsection{Theory of Prior Annealing: Proof of Theorem 3.1}
Our proof follows the proof of Theorem 2 in \cite{chen2015convergence}. SGLD use the first order integrator (see Lemma 12 of \cite{chen2015convergence} for the detail). Then we have 
\begin{equation}
\nonumber
\begin{split}
\mathbb{E}(\psi(\bbeta^{(t+1)})) 
= &\psi(\bbeta^{(t)}) + \epsilon_t \mathcal{L}_t\psi(\bbeta^{(t)}) + O(\epsilon_t^2)\\
=& \psi(\bbeta^{(t)}) + \epsilon_t (\mathcal{L}_t - \mathcal{L})\psi(\bbeta^{(t)}) + \epsilon_t \mathcal{L}\psi(\bbeta^{(t)}) + O(\epsilon_t^2).
\end{split}
\end{equation}
Note that by Poisson equation, $\mathcal{L} \psi(\bbeta) = \phi(\bbeta) - \int \phi(\bbeta) \pi(\bbeta|D_n, \eta^*, \sigma_{0,n}^*)d\bbeta$. Taking expectation on both sides of the equation, summing over $t = 0, 1, \dots, T - 1$, and dividing $\epsilon T$ on both sides of the equation, we have
\begin{equation}
\nonumber
\begin{split}
& \mathbb{E}\left(\frac{1}{T} \sum_{t = 1}^{T - 1} \phi(\bbeta^{(t)}) - \int \phi(\bbeta) \pi(\bbeta|D_n, \eta^*, \sigma_{0,n}^*)\right) \\
= & \frac{1}{T\epsilon}(\mathbb{E}(\psi(\bbeta^{(T)})) - \psi(\bbeta^{(0)})) - \frac{1}{T}\sum_{t = 0}^{T-1} \mathbb{E}(\delta_t \psi(\bbeta^{(t)})) + O(\epsilon).
\end{split}
\end{equation}
To characterize the order of $\delta_t = \mathcal{L}_t - \mathcal{L}$, we first study the difference of the drift term
\begin{equation}
\nonumber
\begin{split}
& \nabla\log(\pi(\bbeta^{(t)}|D_{m,n}^{(t)}, \eta^{(t)}, \sigma_{0,n}^{(t)})) - \nabla\log(\pi(\bbeta^{(t)}|D_n, \eta^*, \sigma_{0,n}^*)) \\
= & \sum_{i=1}^n\nabla\log(p_{\bbeta^{(t)}}(\bx_i, y_i)) - \frac{n}{m} \sum_{j=1}^m\nabla\log(p_{\bbeta^{(t)}}(\bx_{i_j}, y_{i_j})) \\
& +  \eta^{(t)} \nabla\log(\pi(\bbeta^{(t)}| \lambda_n, \sigma_{0,n}^{(t)}, \sigma_{1,n})) - \eta^{*}\nabla \log(\pi(\bbeta^{(t)}| \lambda_n, \sigma_{0,n}^{*}, \sigma_{1,n})).
\end{split}
\end{equation}
Use of the mini-batch data gives an unbiased estimator of the full gradient, i.e. 
\[
\mathbb{E}( \sum_{i=1}^n\nabla\log(p_{\bbeta^{(t)}}(\bx_i, y_i)) - \frac{n}{m} \sum_{j=1}^m\nabla\log(p_{\bbeta^{(t)}}(\bx_{i_j}, y_{i_j}))) = 0.
\]
For the prior part, let $p(\sigma)$ denote the density function of $N(0, \sigma)$. Then we have 
\begin{equation}
\nonumber
\begin{split}
& \nabla\log(\pi(\bbeta^{(t)}| \lambda_n, \sigma_{0,n}^{(t)}, \sigma_{1,n})) \\
= & -\frac{(1-\lambda_n)p(\sigma_{0,n}^{(t)})}{(1-\lambda_n)p(\sigma_{0,n}^{(t)}) + \lambda_n p(\sigma_{1,n})}\frac{\bbeta^{(t)}}{ {\sigma_{0,n}^{(t)}}^2 } -\frac{\lambda_n p(\sigma_{1,n})}{(1-\lambda_n)p(\sigma_{0,n}^{(t)}) + \lambda_n p(\sigma_{1,n})}\frac{\bbeta^{(t)}}{ \sigma_{1,n}^2 },
 \end{split}
\end{equation}
and thus $\mathbb{E}|\nabla\log(\pi(\bbeta^{(t)}| \lambda_n, \sigma_{0,n}^{(t)}, \sigma_{1,n}))| \leq \frac{2\mathbb{E}|\bbeta^{(t)}|}{{\sigma_{0,n}^{*}}^2}$. By Assumption 5.2, we have 
\begin{equation}
\nonumber
\begin{split}
& \mathbb{E}(|\eta^{(t)} \nabla\log(\pi(\bbeta^{(t)}| \lambda_n, \sigma_{0,n}^{(t)}, \sigma_{1,n})) - \eta^{*}\nabla \log(\pi(\bbeta^{(t)}| \lambda_n, \sigma_{0,n}^{*}, \sigma_{1,n}))|) \\
= & \mathbb{E}(|\eta^{(t)} \nabla\log(\pi(\bbeta^{(t)}| \lambda_n, \sigma_{0,n}^{(t)}, \sigma_{1,n})) - \eta^{*}\nabla \log(\pi(\bbeta^{(t)}| \lambda_n, \sigma_{0,n}^{(t)}, \sigma_{1,n}))|)\\
&+ \mathbb{E}(|\eta^{*} \nabla\log(\pi(\bbeta^{(t)}| \lambda_n, \sigma_{0,n}^{(t)}, \sigma_{1,n})) - \eta^{*}\nabla \log(\pi(\bbeta^{(t)}| \lambda_n, \sigma_{0,n}^{*}, \sigma_{1,n}))|) \\
\leq & \frac{2M}{{\sigma_{0,n}^{*}}^2}|\eta^{(t)} - \eta^{*}|  + \eta^{*}M|\sigma_{0,n}^{(t)} - \sigma_{0,n}^{*} |.
\end{split}
\end{equation}
By Assumption 5.1,  $\mathbb{E}(\psi(\bbeta^{(t)})) \leq \infty$. Then   
\[
\frac{1}{T}\sum_{t = 0}^{T-1} \mathbb{E}(\delta_t \psi(\bbeta^{(t)})) = O\left(\frac{1}{T}\sum_{t = 0}^{T-1} (|\eta^{(t)} - \eta^*| +|\sigma_{0,n}^{(t)} - \sigma_{0,n}^{*} |) \right).
\]
Note that by assumption 5.1, $|(\psi(\bbeta^{(T)})) - \psi(\bbeta^{(0)})|$ is bounded. Then 
\[
\mathbb{E}\left(\frac{1}{T} \sum_{t = 1}^{T - 1} \phi(X_t) - \int \phi(\bbeta) \pi(\bbeta|D_n, \eta^*, \sigma_{0,n}^*) \right) = O\left(\frac{1}{T\epsilon} + \frac{\sum_{t = 0}^{T-1} (|\eta^{(t)} - \eta^*| +|\sigma_{0,n}^{(t)} - \sigma_{0,n}^{*} |)}{T} + \epsilon\right).
\]

\subsection{Construct Confidence Interval}

Theorem 2.2 implies that a faithful prediction interval can be constructed for the sparse neural network learned by the proposed algorithms. In practice, for a normal regression problem with noise $N(0,\sigma^2)$, to construct the prediction interval for a test point $\bx_0$, the terms $\sigma^2$ and $\Sigma = \nabla_{\bgamma^*}\mu(\bbeta^*, \bx_0)^{T}H^{-1}\nabla_{\bgamma^*}\mu(\bbeta^*, \bx_0)$ in Theorem 2.2 need to be estimated from data. 
Let $D_{n}=(\bx^{(i)},y^{(i)})_{i=1,...,n}$ be the training set and $\mu(\bbeta, \cdot)$ be the predictor of the network model with parameter $\bbeta$. We can follow the following procedure to construct the prediction interval for the  test point $\bx_0$:
\begin{itemize}
\item Run algorithm 1, let $\hat{\bbeta}$ be an estimation of the network parameter at the end of the algorithm and $\hat{\bgamma}$ be the correspoding network structure. 
\item Estimate $\sigma^2$ by 
\[
\hat{\sigma}^2 = \frac{1}{n} \sum_{i=1}^n ( \mu(\hat{\bbeta}, \bx^{(i)}) - y^{(i)})^2.
\] 
\item Estimate $\Sigma$ by 
\[
\hat{\Sigma} =  \nabla_{\hat{\bgamma}}\mu(\hat{\bbeta}, \bx_0)^{T}(-\nabla^2_{\hat{\bgamma}} l_n(\hat{\bbeta}))^{-1}\nabla_{\hat{\bgamma}}\mu(\hat{\bbeta}, \bx_0).
\]
\item Construct the prediction interval as 
\[
\left(\mu(\hat{\bbeta}, \bx_0) - 1.96\sqrt{\frac{1}{n}\hat{\Sigma} + \hat{\sigma}^2}, \mu(\hat{\bbeta}, \bx_0) + 1.96\sqrt{\frac{1}{n}\hat{\Sigma} + \hat{\sigma}^2}\right).
\]
\end{itemize}
Here, by the structure selection consistency (Lemma 2.2) and consistency of the MLE for the learnt structure \cite{Portnoy1988}, we replace $\bbeta^*$ and $\bgamma^*$ in Theorem 2.2 by $\hat{\bbeta}$ and $\hat{\bgamma}$.

If the dimension of the sparse network is still too high and the computation of $\hat{\Sigma}$ becomes prohibitive, the following Bayesian approach can be used to construct confidence intervals. 
\begin{itemize}
    \item Running SGMCMC algorithm to get a sequence of posterior samples: $\bbeta^{(1)},\dots, \bbeta^{(m)}$.
    \item Estimating $\sigma^2$ by $ \hat{\sigma}^2 = \frac{1}{n}\sum_{i=1}^n(y^{(i)} - \mu^{(i)})^2$, where 
    \begin{equation}  \nonumber
    \mu^{(i)} = \frac{1}{m}\sum_{j=1}^m\mu(\bbeta^{(j)}, \bx^{(i)}), i = 1,\dots, n,
    \end{equation}
    \item Estimate the prediction mean by 
    \[
    \hat{\mu} = \frac{1}{m} \sum_{i=1}^{m} \mu(\bbeta^{(i)}, \bx_0).
    \]
    \item Estimate the prediction variance by
    \[
\hat{V} = \frac{1}{m} \sum_{i=1}^{m}(\mu(\bbeta^{(i)},\bx_0) - \hat{\mu})^2 + \hat{\sigma}^2.
    \]
    \item Construct the prediction interval as
    \[
    (\mu-1.96\sqrt{V}, \mu+1.96\sqrt{V}).
    \]
\end{itemize}

\subsection{Prior Annealing}

\begin{figure}
\centering
\includegraphics[scale = 0.4]{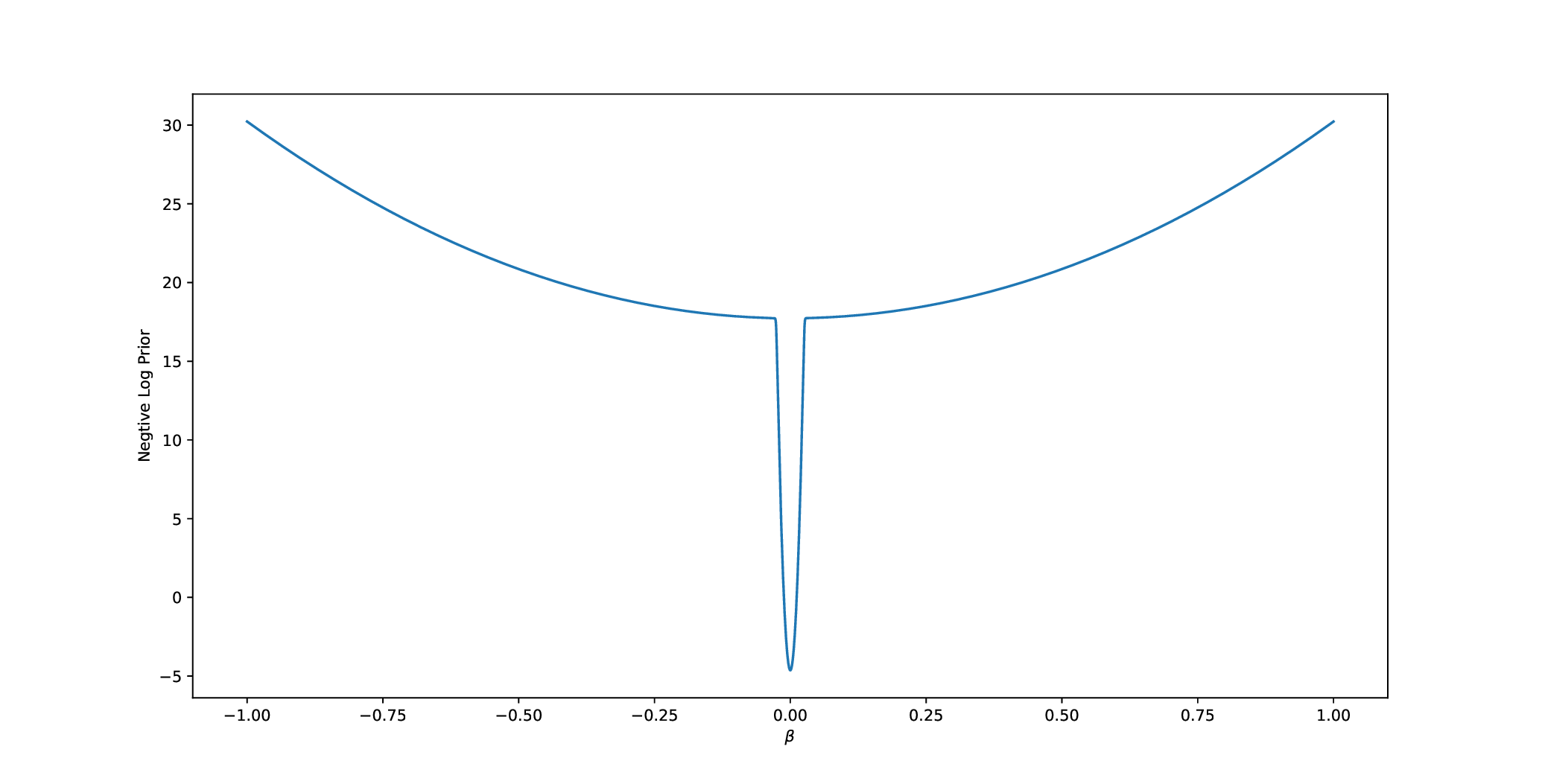}
\caption{Negative logarithm of the mixture Gaussian prior.}
\label{negative_log_prior}
\vspace{-0.1in}
\end{figure}

In this section, we give some graphical illustration of the prior annealing algorithm. In practice, the negative log-prior puts penalty on parameter weights. The mixture Gaussian prior behaves like a piecewise $L_2$ penalty with different weights on different regions. Figure \ref{negative_log_prior} shows the shape of a negative log-mixture Gaussian prior. In step (iii) of Algorithm 1, the condition $\pi(\bgamma_i = 1 | \bbeta_i) > 0.5$ splits the parameters into two parts. For the $\bbeta_i$'s with large magnitudes, the slab component $N(0, \sigma_{1,n}^2)$ plays the major role in the prior, imposing a small penalty on the parameter. For the $\bbeta_i$'s with smaller magnitudes, the spike component $N(0, \sigma_{0,n}^2)$ plays the major role in the prior, imposing a large penalty on the parameters to push them toward zero in training. 

\begin{figure}
\centering
\begin{tabular}{c}
\includegraphics[scale = 0.4]{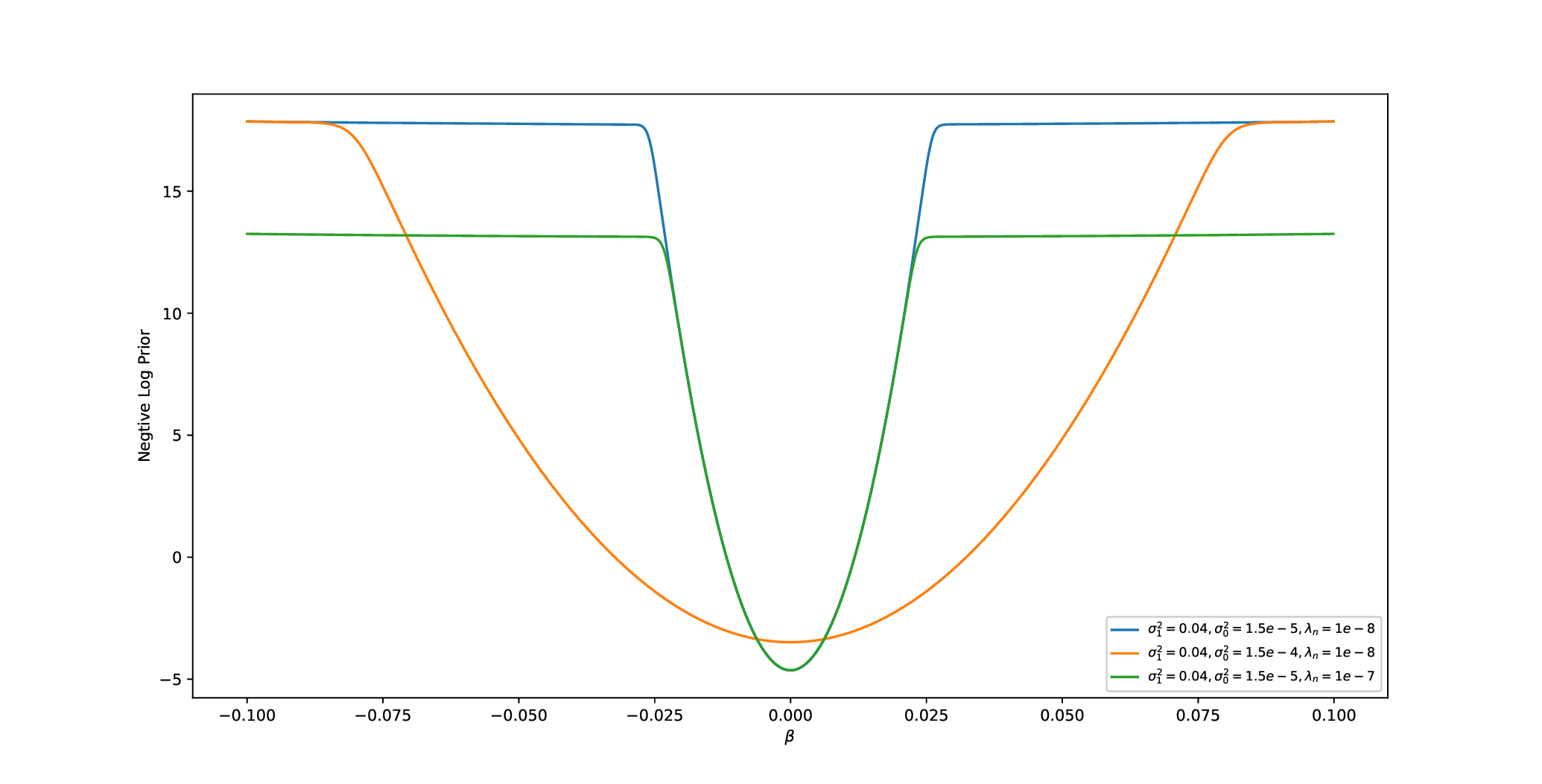} \\
\includegraphics[scale = 0.4]{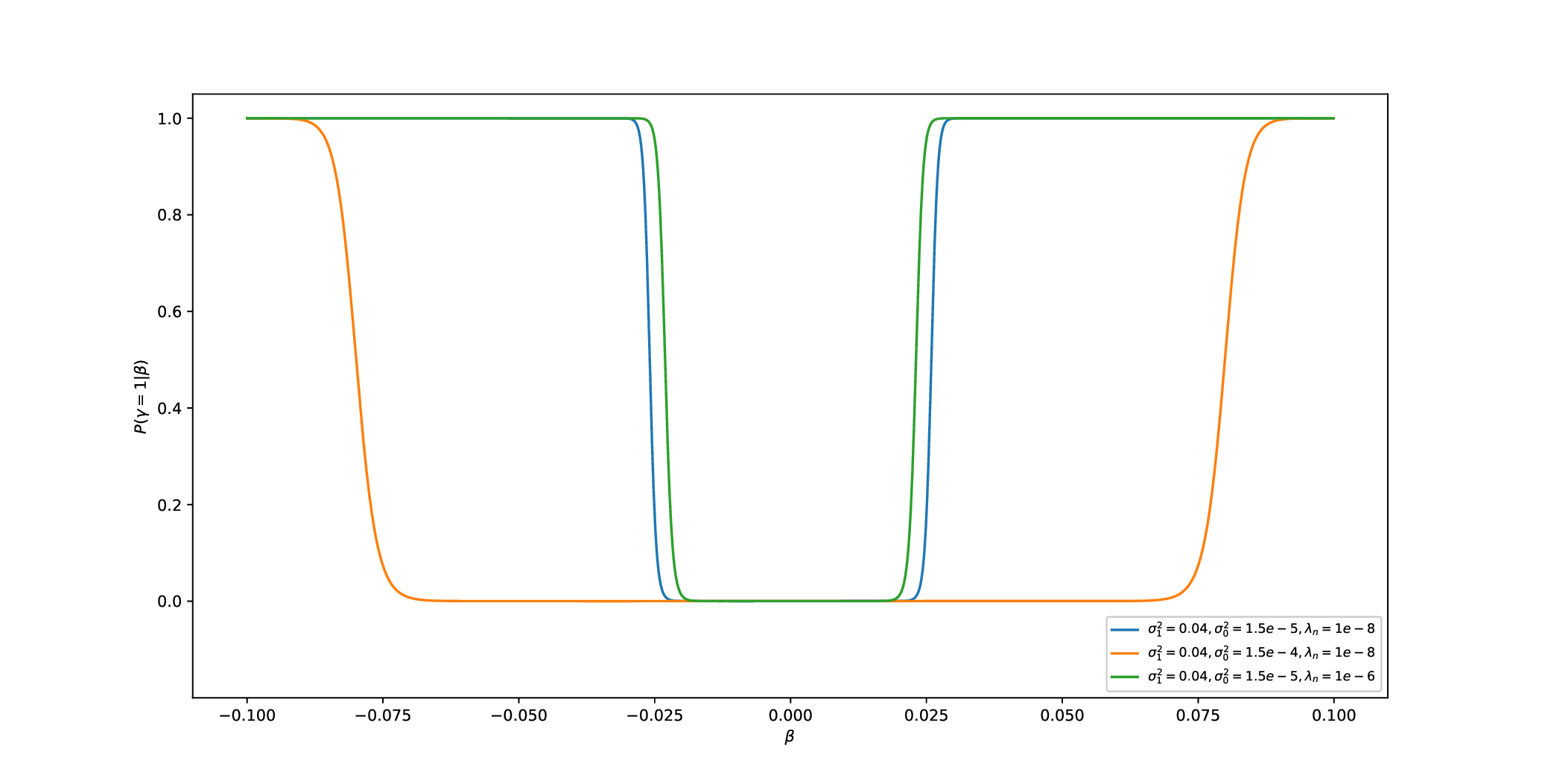}
\end{tabular}
\caption{Negative log-prior and $\pi(\bgamma = 1 | \bbeta)$ for different choices of $\sigma_{0,n}^2$ and $\lambda_n$.}
\label{prior_anneal}
\vspace{-0.1in}
\end{figure}

Figure \ref{prior_anneal} shows the shape of negative log-prior and $\pi(\bgamma_i = 1| \bbeta_i )$ for different choices of $\sigma_{0,n}^2$ and $\lambda_n$. As we can see from the plot,  $\sigma_{0,n}^2$ plays the major role in determining the effect of the prior. Let $\alpha$ be the threshold in step (iii) of Algorithm 1 of the main body, i.e. the positive solution to $\pi(\bgamma_i = 1 | \bbeta_i) = 0.5$. In general, a smaller $\sigma_{0,n}^2$ will result in a smaller $\alpha$. If a very small $\sigma_{0,n}^2$ is used in the prior from the beginning, then most of $\bbeta_i$'s at initialization will have a magnitude larger than $\alpha$ and the slab component $N(0, \sigma_{1,n}^2)$ of the prior will dominate most parameters. As a result, it will be difficult to find the desired sparse structure. Following the proposed prior annealing procedure, we can start with a larger $\sigma_{0,n}^2$, i.e. a larger threshold $\alpha$ and a relatively smaller penalty for those $|\bbeta_i| < \alpha$. As we gradually decrease the value of $\sigma_{0,n}^2$, $\alpha$ decreases, and the penalty imposed on the small weights increases and drives them toward zero. 
The prior annealing allows us to gradually sparsify the DNN and impose more and more penalties on the parameters close to 0. 

\subsection{Experimental Setups}

\subsubsection{Simulated examples}

\paragraph{Prior annealing} 
We follow simple implementation of Algorithm given in section 3.1.
We run SGHMC for $T = 80000$ iterations with constant learning rate $\epsilon_t = 0.001$,  momentum $1-\alpha = 0.9$ 
and subsample size $m = 500$. We set $\lambda_n = 1e-7, \sigma_{1,n}^2 = 1e-2$, 
$(\sigma_{0,n}^{init})^2 = 5e-5$, $(\sigma_{0,n}^{end})^2 = 1e-6$ and $T_1 = 5000, T_2 = 20000, T_3 = 60000$. We set temperature $\tau = 0.1$ for $t < T_3$ and for $t > T_3$, we gradually decrease temperature $\tau$ by $\tau = \frac{0.1}{t - T_3}$. After structure selection, the model is fine tuned for 40000 iterations. The number of iteration setup is the same as \cite{SunSLiang2021}.

\paragraph{Other Methods}
Spinn, Dropout and DNN are trained with the same network structure using SGD with momentum. Same as our method, we use constant learning rate $0.001$, momentum $0.9$, subsample size $500$ and traing the model for 80000 iterations. For Spinn, we use LASSO penalty and the regularization parameter is selected from $\{0.05, 0.06, \dots, 0.15\}$ according to the performance on validation data set. For Dropout, the dropout rate is set to be $0.2$ for the first layer and $0.5$ for the other layers. Other baseline methods BART50, LASSO, SIS are implemented using R-package $randomForest$, $glmnet$, $BART$ and $SIS$ respectively with default parameters.

\subsubsection{CIFAR10}

 We follow the standard training procedure as in \cite{lin2020dynamic}, i.e. we train the model with SGHMC for $T = 300$ epochs, with initial learning rate $\epsilon_0 = 0.1$, momentum $1-\alpha = 0.9$, temperature $\tau = 0.001$, mini-batch size $m = 128$. The learning rate is divided by 10 at $150$th and $225$th epoch. We follow the implementation given in section 3.1 and use $T_1 = 150, T_2 = 200, T_3 = 225$, where $T_i$s are number of epochs. We set temperature $\tau = 0.01$ for $t < T_3$ and gradually decrease $\tau$ by $\tau = \frac{0.01}{t - T_3}$ for $t > T_3$. We set $\sigma_{1,n}^2 = 0.04$ and $(\sigma_{0,n}^{init})^2 = 10 \times (\sigma_{0,n}^{end})^2$ and use different $\sigma_{0,n}^{end}, \lambda_n$ for different network size and target sparsity level. The detailed settings are given below:
\begin{itemize}
\item ResNet20 with target sparsity level 20\%: $(\sigma_{0,n}^{end})^2 = 1.5e-5,  \lambda_n = 1e-8$
\item ResNet20 with target sparsity level 10\%: $(\sigma_{0,n}^{end})^2 = 6e-5,  \lambda_n = 1e-9$
\item ResNet32 with target sparsity level 10\%: $(\sigma_{0,n}^{end})^2 = 3e-5,  \lambda_n = 2e-9$
\item ResNet32 with target sparsity level 5\%: $(\sigma_{0,n}^{end})^2 = 1e-4,  \lambda_n = 2e-8$

\end{itemize}

\end{document}